\documentclass[10pt,journal,compsoc]{IEEEtran}
\ifCLASSOPTIONcompsoc
  \usepackage[nocompress]{cite}
\else
  \usepackage{cite}
\fi
\ifCLASSINFOpdf
\else
\fi

\usepackage{footnote}
\usepackage{url}

\usepackage{amsmath}
\usepackage{algorithmicx}
\usepackage{algorithm}
\usepackage{amssymb}
\usepackage{graphicx}
\usepackage{subfigure}
\usepackage{multirow}
\usepackage{booktabs}
\usepackage{bbm}
\usepackage{bm}
\usepackage{algpseudocode}
\usepackage{arydshln}
\usepackage{xcolor}

\begin{document}
%
\title{A Unified Contrastive Transfer Framework with Propagation Structure for Boosting Low-Resource Rumor Detection}

\author{Hongzhan~Lin,
        Jing~Ma,
        Ruichao~Yang,
        Zhiwei~Yang,
        and~Mingfei~Cheng
\IEEEcompsocitemizethanks{\IEEEcompsocthanksitem Hongzhan Lin, Jing Ma, Ruichao Yang are with the Department
of Computer Science, Hong Kong Baptist University, Hong Kong.\protect\\
E-mail: \{cshzlin, majing, csrcyang\}@comp.hkbu.edu.hk
\IEEEcompsocthanksitem Zhiwei Yang is with Jilin University, China.\protect\\
E-mail: yangzw18@mails.jlu.edu.cn
\IEEEcompsocthanksitem Mingfei Cheng is with Singapore Management University, Singapore.\protect\\
E-mail: mfcheng.2022@phdcs.smu.edu.sg}
\thanks{Jing Ma is the corresponding author.}}

%
%

\ifCLASSOPTIONpeerreview
\markboth{Journal of \LaTeX\ Class Files,~Vol.~14, No.~8, August~2015}%
{Lin \MakeLowercase{\textit{et al.}}: A Unified Contrastive Transfer Framework with Propagation Structure for Boosting Low-Resource Rumor Detection}
%
\fi



\IEEEtitleabstractindextext{%
\begin{abstract}
The truth is significantly hampered by massive rumors that spread along with breaking news or popular topics. Since there is sufficient corpus gathered from the same domain for model training, existing rumor detection algorithms show promising performance on yesterday's news. However, due to a lack of substantial training data and prior expert knowledge, they are poor at spotting rumors concerning unforeseen events, especially those propagated in different languages (i.e., low-resource regimes). In this paper, we propose a plug-and-play framework with unified contrastive transfer learning, to detect rumors by adapting the features learned from well-resourced rumor data to that of the low-resourced with only few-shot annotations. More specifically, we first represent rumor circulated on social media as an undirected topology for enhancing the interaction of user opinions, and then train the propagation-structured model via a unified contrastive paradigm to mine effective clues simultaneously from both propagation structure and post semantics. 
Our model explicitly breaks the barriers of the domain and/or language issues, via language alignment and a novel domain-adaptive contrastive learning mechanism. To well-generalize the representation learning using a small set of annotated target events, we reveal that rumor-indicative signal is closely correlated with the uniformity of the distribution of these events. We design a target-wise contrastive training mechanism with three event-level data augmentation strategies, capable of unifying the representations by distinguishing target events. 
Extensive experiments conducted on four low-resource datasets collected from real-world microblog platforms demonstrate that our simple yet effective framework achieves much better performance than state-of-the-art methods and exhibits a superior capacity for detecting rumors at early stages.
\end{abstract}

}

\maketitle

\IEEEdisplaynontitleabstractindextext

%
\IEEEpeerreviewmaketitle

\IEEEraisesectionheading{\section{Introduction}\label{sec:introduction}}
\IEEEPARstart{R}umor amid breaking news spreads like wildfire on social media, causing widespread confusion, fear, and distrust among individuals and society. However, rumor detection can be particularly challenging in low-resourced domains or languages due to the availability of domain expertise, nuances of language, cultural differences, etc.
Taking the healthcare domain as an example, during the outbreak of COVID-19, 
a false rumor claimed ``the vaccine has a chip in it which will control your mind"\footnote{\url{https://www.bbc.com/news/55768656}}. 
The rumor was translated into various languages, enabling it to spread in different regions worldwide, including those with low levels of vaccine uptake or hesitant attitudes towards vaccination like Arabic, India, and other Muslim countries.
Despite recent efforts to collect microblog posts related to COVID-19~\cite{chen2020covid, zarei2020first, alqurashi2020large}, existing rumor detection methods are vulnerable to detecting 
such low-resource rumors without a substantial and suitable training corpus~\cite{hedderich2021survey}. Therefore, to mitigate their harmful effects, it is crucial to develop robust methods to detect rumors in low-resource domains and languages during breaking news events.

{
\begin{figure*}[ht]
\centering
\subfigure[TWITTER (Rumor)]{
\begin{minipage}[t]{0.33\linewidth}
\centering
\scalebox{0.85}{\includegraphics[width=6cm]{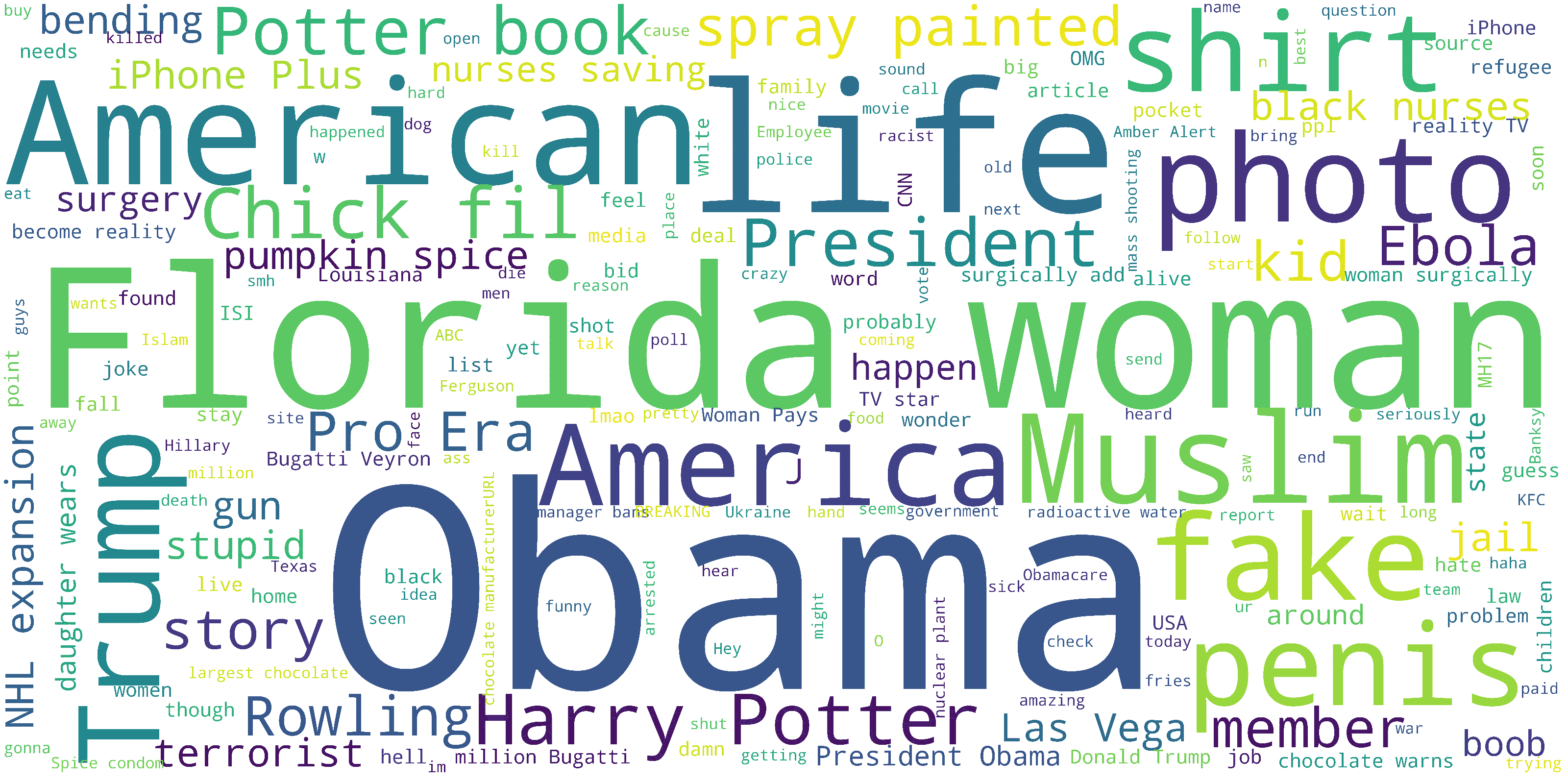}}
\label{fig:motivation_a}
\end{minipage}%
}%
\subfigure[English-COVID19 (Rumor)]{
\begin{minipage}[t]{0.33\linewidth}
\centering
\scalebox{0.85}{\includegraphics[width=6cm]{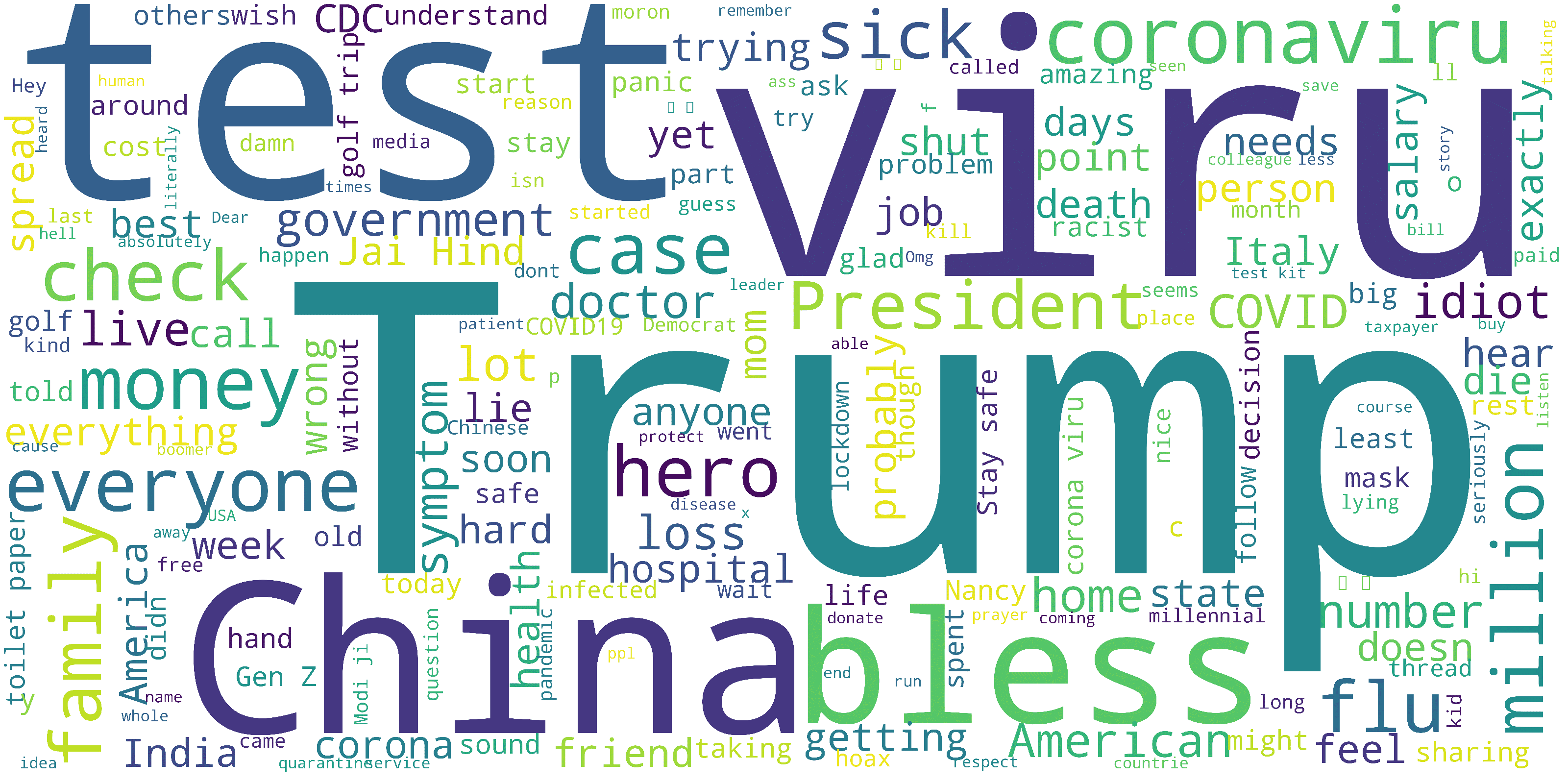}}
\label{fig:motivation_b}
\end{minipage}%
}%
\subfigure[Chinese-COVID19 (Rumor)]{
\begin{minipage}[t]{0.33\linewidth}
\centering
\scalebox{0.85}{\includegraphics[width=6cm]{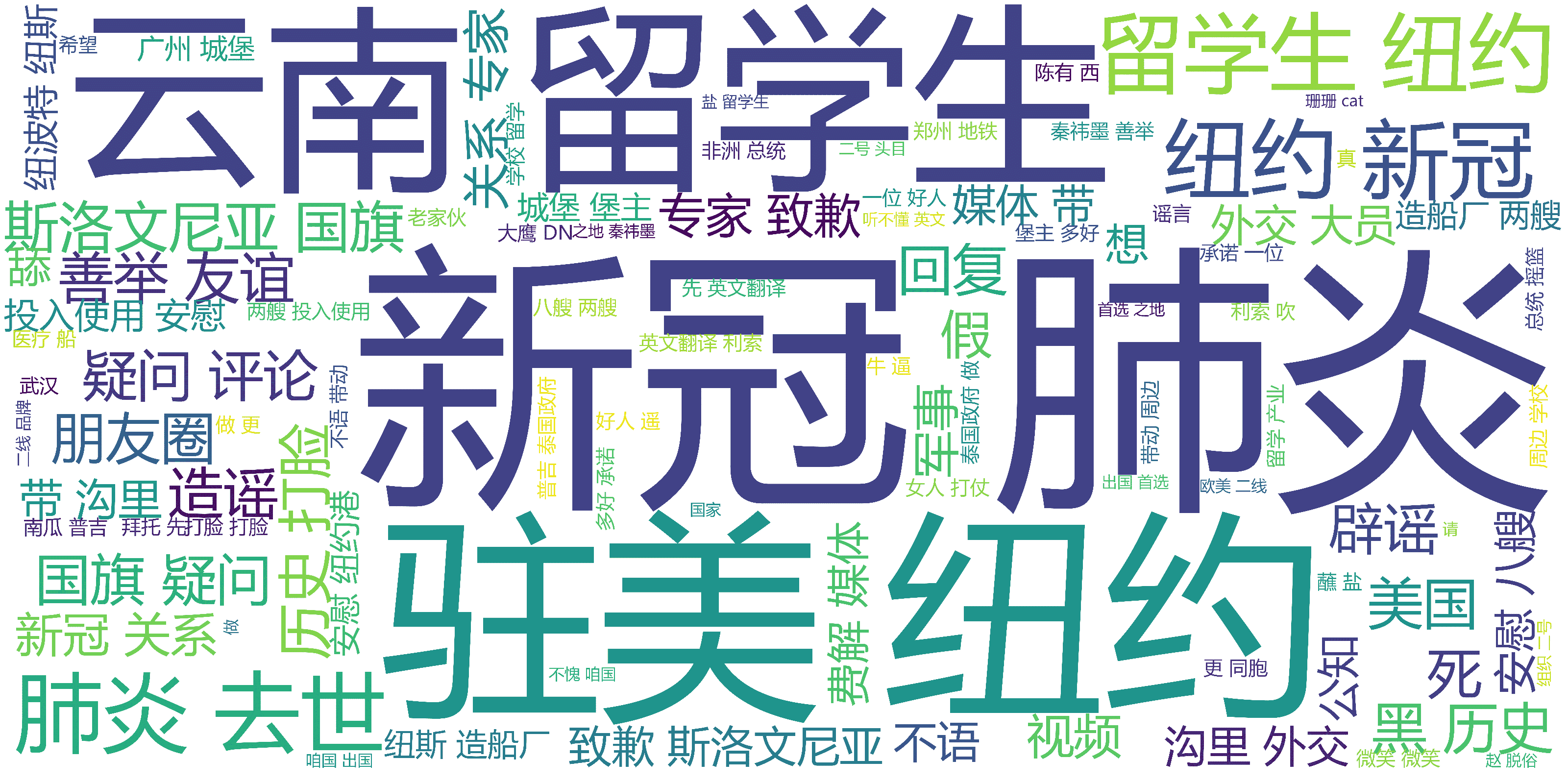}}
\label{fig:motivation_c}
\end{minipage}%
}%
\\
\subfigure[TWITTER (Non-rumor)]{
\begin{minipage}[t]{0.33\linewidth}
\centering
\scalebox{0.85}{\includegraphics[width=6cm]{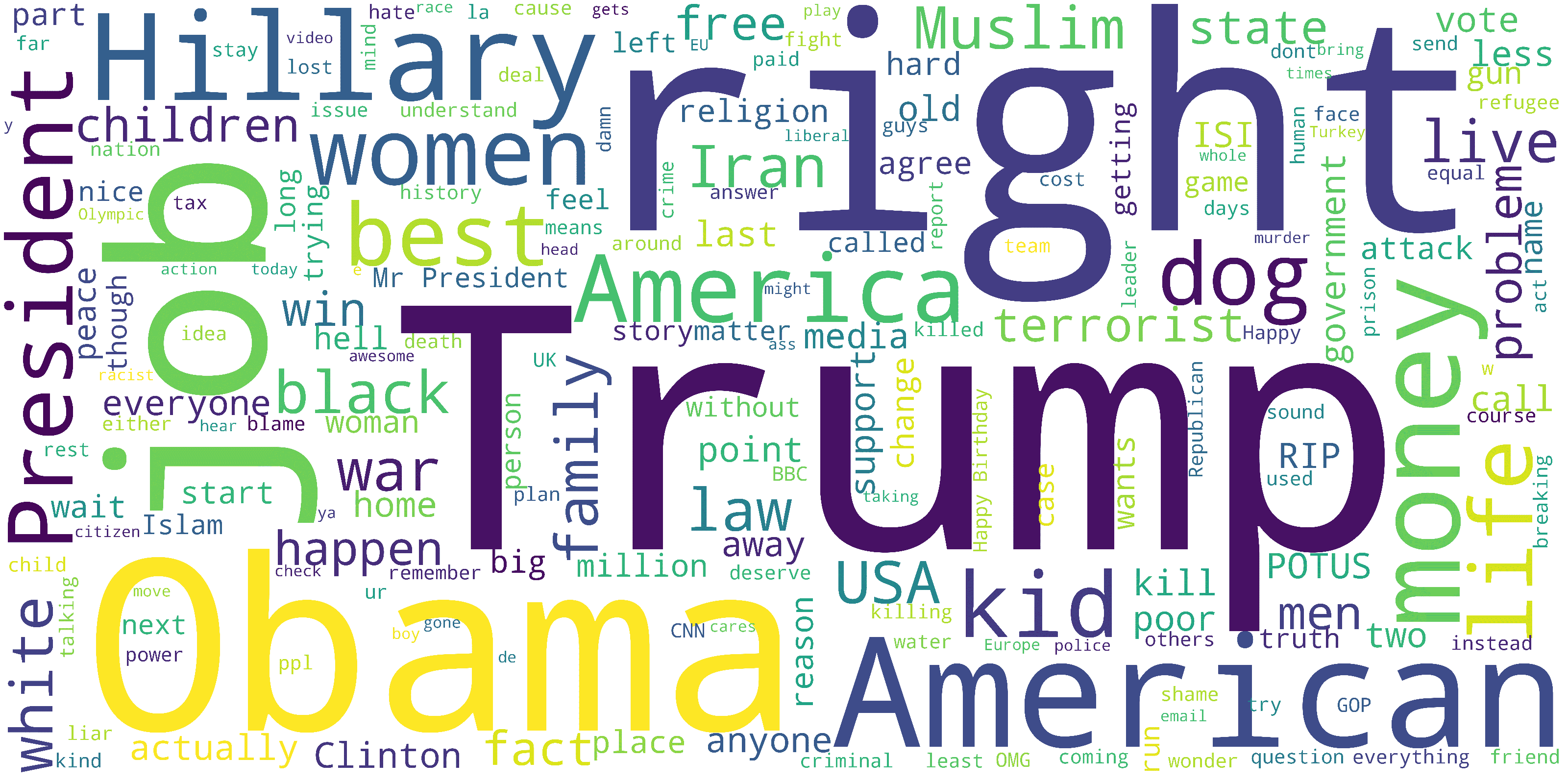}}
\label{fig:motivation_d}
\end{minipage}%
}%
\subfigure[English-COVID19 (Non-rumor)]{
\begin{minipage}[t]{0.33\linewidth}
\centering
\scalebox{0.85}{\includegraphics[width=6cm]{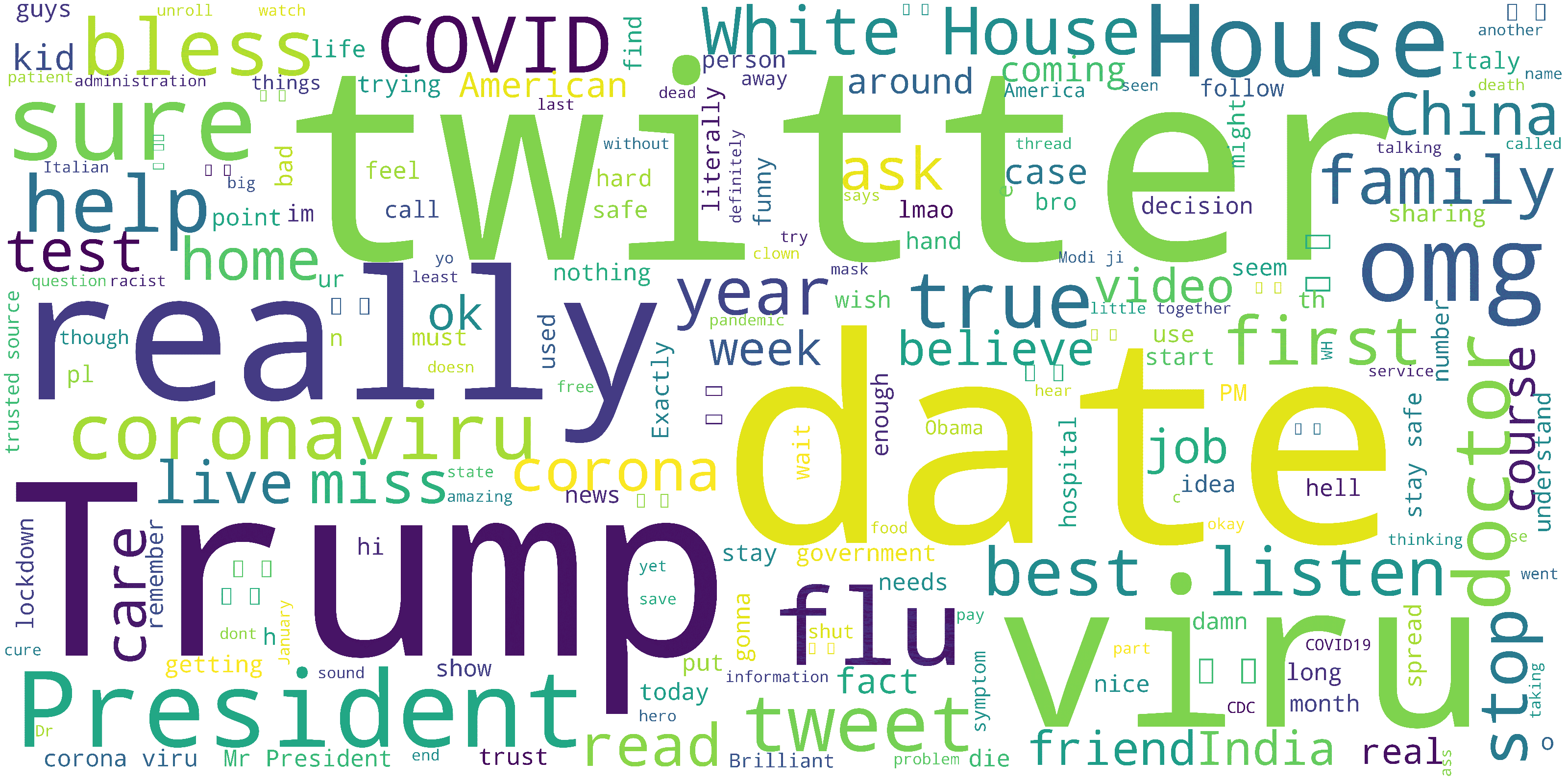}}
\label{fig:motivation_e}
\end{minipage}%
}%
\subfigure[Chinese-COVID19 (Non-rumor)]{
\begin{minipage}[t]{0.33\linewidth}
\centering
\scalebox{0.85}{\includegraphics[width=6cm]{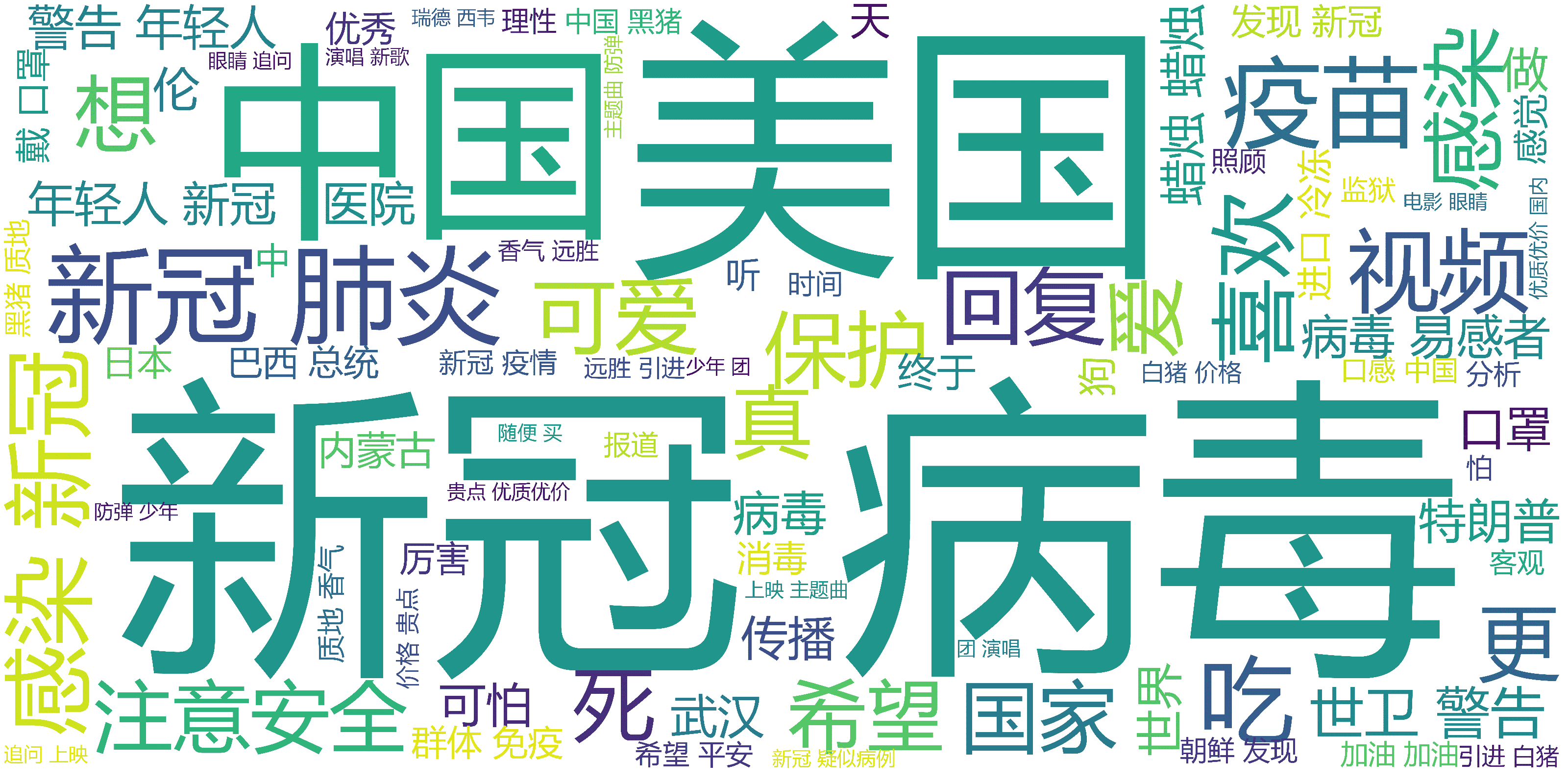}}
\label{fig:motivation_f}
\end{minipage}%
}%
\centering
\caption{Word clouds of rumor and non-rumor data generated from TWITTER, English-COVID19, and Chinese-COVID19 datasets, where the size of terms corresponds to the word frequency. Both TWITTER and English-COVID19 are presented in English while Chinese-COVID19 in Chinese.}
\label{fig:motivation}
\end{figure*}}

Social psychology literature defines a rumor as a story or a statement whose truth value is unverified or deliberately false~\cite{allport1947psychology}. 
Recently, deep neural network (DNN) techniques~\cite{ma2018rumor, khoo2020interpretable, bian2020rumor} have shown great potential in detecting rumors on microblogging websites by extracting features indicative of rumors from a sizeable corpus of labeled rumor data. Nevertheless, such DNN-based approaches are purely data-driven and have a major limitation on detecting emerging events concerning about low-resource domains, i.e., the distinctive topic coverage and word distribution~\cite{silva2021embracing} required for detecting low-resource rumors are often not covered by the public benchmarks~\cite{zubiaga2016learning, ma2016detecting, ma2017detect}. On another hand, for rumors propagated in different languages, existing monolingual approaches are not applicable since there are even no sufficient open domain data for model training in the target language.

In this study, we assume that establishing close correlations between well-resourced and low-resourced rumor data can help overcome domain and language barriers, thereby improving low-resource rumor detection within a more comprehensive framework. 
To illustrate our intuition, we collect rumorous and non-rumorous claims corresponding to COVID-19 with propagation threads from Twitter and Sina Weibo which are two popular social websites in English-spoken and Chinese-spoken communities, respectively. Figure~\ref{fig:motivation} illustrates the word clouds of rumor and non-rumor data from an open domain benchmark (i.e., TWITTER ~\cite{ma2017detect}) and two COVID-19 datasets~\cite{lin2022detect} (i.e., English-COVID19 and Chinese-COVID19). It can be seen that both TWITTER and English-COVID19 contain denial opinions towards rumors, e.g., ``fake", ``joke", ``stupid" in Figure~\ref{fig:motivation_a} and ``wrong symptom", ``exactly sick", ``health panic" in Figure~\ref{fig:motivation_b}. In contrast, supportive opinions towards non-rumors can be drawn from Figure~\ref{fig:motivation_d}--\ref{fig:motivation_e}. 
Moreover, considering that COVID-19 is a global disease, massive misinformation could be widely propagated in different languages such as Arabic~\cite{alam2021fighting}, Indic~\cite{kar2021no}, English~\cite{cui2020coaid} and Chinese~\cite{hu2020weibo}. Similar identical patterns can be observed in Chinese on Sina Weibo from Figure~\ref{fig:motivation_c} and Figure~\ref{fig:motivation_f}.
Although the COVID-19 data tend to use domain-specific expertise jargon or language-related colloquialisms on social media, we argue that aligning the representation space of identical rumor-indicative patterns of different domains and/or languages can help adapt the features captured from well-resourced data to that of the low-resourced. Moreover, since rumor propagation generally reveals significant insight into how a claim is responded to by users irrespective of specific domains~\cite{zubiaga2018detection, ma2018rumor}, we aim to develop an innovative domain and/or language transfer framework that is aware of such a structural social context.

To this end, inspired by contrastive learning~\cite{he2020momentum, chen2020simple, chen2020big}, we proposed a domain-Adaptive Contrastive Learning approach for low-resource rumor detection, to encourage effective alignment of rumor-indicative features in the well-resourced and low-resource data. More specifically, we first transform each microblog post into a language-independent vector by semantically aligning the source and target language in a shared vector space. As the diffusion of rumors generally follows a propagation structure that provides valuable domain-invariant clues on how a claim is transmitted, we present the conversation propagation thread as an undirected topology, which allows full-duplex interactions between posts with responsive relationships so that the domain-invariant structural features can be fully aggregated and the interplay of user viewpoints can be enhanced. Thus we resort to a multi-scale Graph Convolutional mechanism to catch informative patterns fused from both claim semantics and event structure. Then, we propose a novel domain-adaptive contrastive learning paradigm to minimize the intra-class variance of source and target instances with same veracity, and maximize inter-class variance of instances with different veracity. 

{
\begin{figure}[t]
    \begin{center}
    \setlength{\belowcaptionskip}{0.1cm}
    \resizebox{0.5\textwidth}{!}{\includegraphics{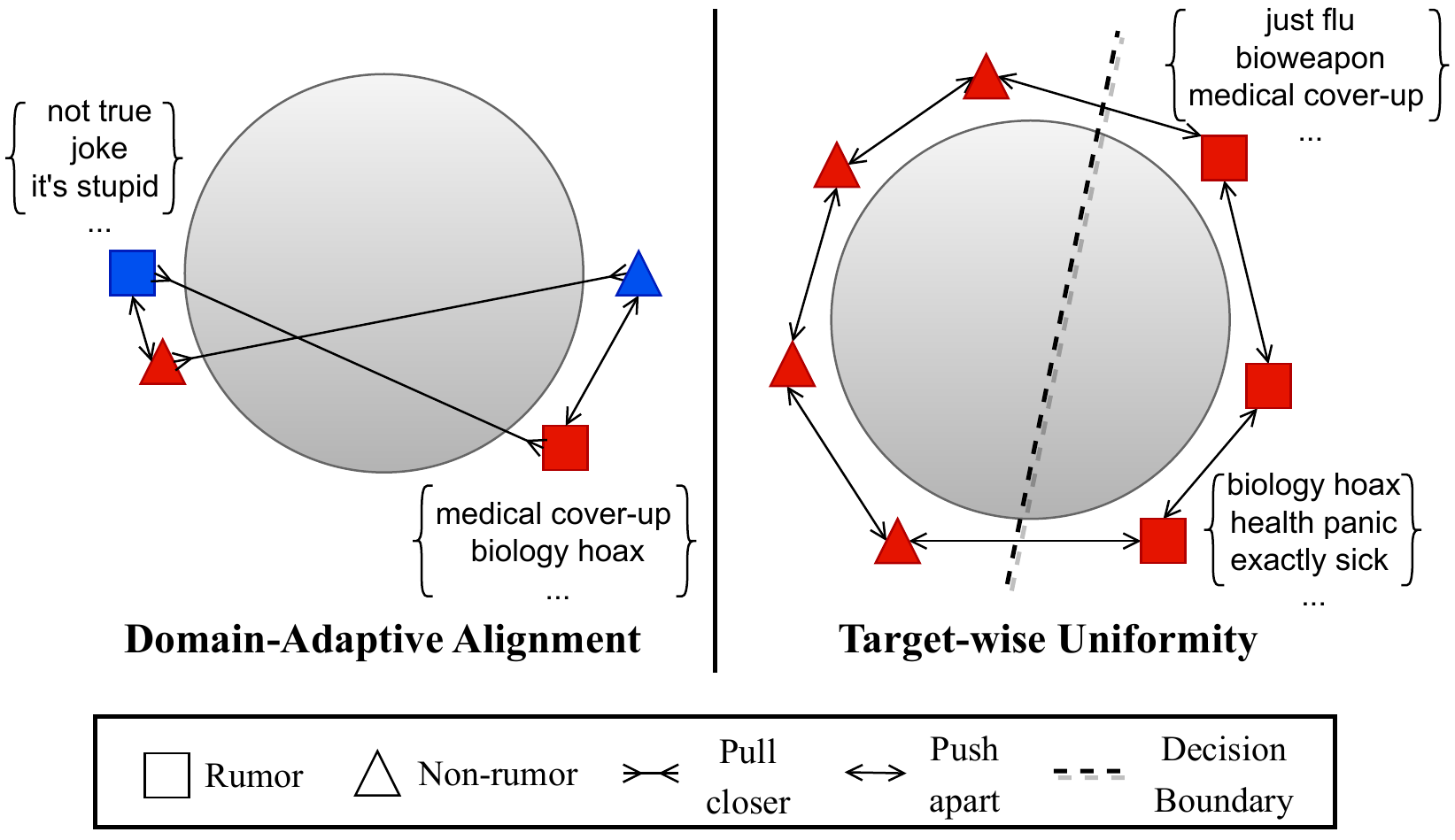}}
    \end{center}
    \caption{The illustration of Alignment and Uniformity of features in source data (in blue) and target data (in red) on a hypersphere. The alignment dedicates identical rumor-indicative features from different domains closer, while the uniformity could help preserve maximal information of the features from target rumor data to capture nontrivial but more discriminative patterns for better generalization. 
    }
    \label{fig:align_uniform}
\end{figure}}

Previous literature reveals that two properties: 1) alignment and 2) uniformity on the unit hypersphere~\cite{wang2020understanding}, are of great importance for representation learning in terms of contrastive paradigm. However, a problem of our domain-adaptive contrastive learning framework proposed in~\cite{lin2022detect} is that it primarily emphasizes the alignment of different domains and/or languages, while largely disregarding the uniformity of target feature space that preserves distinctive rumor-indicative signals among target training samples~\cite{bachman2019learning}. The inductive bias of such an alignment-only paradigm could cause the representation degeneration issue~\cite{gaorepresentation} and limit the capacity of generalization to unlabeled low-resource rumor data.
Figure~\ref{fig:align_uniform} illustrates the alignment between the source rumor data in open domains and a small amount of labeled target data in COVID-19 domain, and the ideal uniformity of target representation learning. The domain-adaptive contrastive learning can align the target rumor data containing target-specific patterns like ``medical cover-up" or ``biology hoax", with the well-resourced data containing general rumor-indicative patterns like ``not true" or ``it's stupid". So that almost all the few-shot target rumor data with the same veracity are shrinking towards the source data and thus degenerating into a narrow and anisotropic feature space, which however suppresses the uniformity of target representation learning. Generally, the more uniform the distribution of the small labeled target data, the more rumor-indicative information is retained, which can lead to better generalization to more target low-resource data~\cite{bachman2019learning}. As exemplified in Figure~\ref{fig:align_uniform}, with an evenly distributed representation on the unit hypersphere~\cite{tian2020contrastive}, the target vector space is able to preserve enough discriminative information related to medicine or biology for each sample, which can be well-generalized on detecting COVID-19 rumors with nontrivial domain-specific patterns like ``just flu" or ``it's bioweapon". 

%
In this work, we introduce a novel unified contrastive transfer framework incorporating target-wise contrastive learning to enhance our basic domain-adaptive model for low-resource rumor detection. The framework aims to improve the generalization to more challenging target events by unifying the few-shot target representation. Meanwhile, in target-wise contrastive learning, data augmentation is a crucial aspect as it enables the creation of new positive and negative pairs from a small amount of labeled target data. This increases the amount and diversity of target training data, improving the performance of detecting more challenging low-resource rumors.
%
Therefore, we explore three event-level data augmentation strategies for target low-resource data (i.e., Adversarial Attack~\cite{goodfellow2014generative, kurakin2016adversarial}, Feature Dropout~\cite{hinton2012improving}, and Graph Dropedge~\cite{rongdropedge}) to effectively obtain pseudo diverse views of a target event in the latent space for the target-wise contrastive learning.

As there is no public benchmark available for detecting low-resource rumors with conversation threads, we collected the propagation structure for four rumor datasets corresponding to COVID-19 from social media in Mandarin, English, Cantonese and Arabic languages. Extensive experiments conducted on four real-world low-resource datasets confirm that (1) our model yields outstanding performances for detecting low-resource rumors over the state-of-the-art baselines with a large margin; (2) the unified contrastive transfer framework is more effective in contrastive representation learning for uniformity of target data distribution;
and (3) our method performs particularly well on early rumor detection which is crucial for timely intervention and debunking especially for breaking events.

The main contributions of this paper are four-fold:
\begin{itemize}
\item To our best knowledge, we are the first to study low-resource rumor detection on social media, by presenting a simple yet effective plug-and-play framework via unified contrastive transfer with propagation structure.  
\item We propose a contrastive learning framework for structural feature adaption between different domains and languages, which model domain-invariant similarities based on undirected propagation topology by pulling together events of the same veracity while pushing apart events of different veracity.

\item Based on the domain-adaptive model previously proposed in our recent work~\cite{lin2022detect}, we design a target-wise contrastive learning mechanism with three event-level data augmentation strategies, to uniform the distributed event-level representations for the target events, capable of learning target-specific rumor signals.
\item We collect four low-resource rumor benchmarks corresponding to COVID-19 domain with conversation threads, respectively represented in Chinese, English, Cantonese and Arabic languages. 
Experimental results on the four real-world benchmarks show that our model achieves superior performance for both rumor classification and early detection tasks under low-resource settings.
\end{itemize}


\section{Related Work}
Pioneer studies for automatic rumor detection focus on learning a supervised classifier utilizing features crafted from post contents, user profiles, and propagation patterns~\cite{castillo2011information, yang2012automatic, liu2015real}. Subsequent studies then propose new features such as those representing rumor diffusion and cascades \cite{kwon2013prominent, friggeri2014rumor, hannak2014get}. \cite{zhao2015enquiring} alleviate the engineering effort by using a set of regular expressions to find questing and denying tweets. DNN-based models such as recurrent neural networks~\cite{ma2016detecting}, convolutional neural networks~\cite{yu2017convolutional}, and attention mechanism~\cite{guo2018rumor} are then employed to learn the features from the stream of social media posts. However, these approaches simply model the post structure as a sequence while ignoring the complex propagation structure. 

To extract useful clues jointly from content semantics and propagation structures, some approaches propose kernel-learning models~\cite{wu2015false, ma2017detect} to make a comparison between propagation trees. Tree-structured recursive neural networks (RvNN)~\cite{ma2018rumor} and transformer-based models~\cite{khoo2020interpretable, ma2020debunking} are proposed to generate the representation of each post along a propagation tree guided by the tree structure. 
More recently, graph neural networks (GNN)~\cite{bian2020rumor, lin2021rumor} have been exploited to encode the conversation thread for higher-level representations. 
However, such data-driven approaches fail to detect rumors in low-resource regimes~\cite{janicka2019cross} because they often require sizeable training data that is not available for low-resource domains and/or languages. In this paper, we propose a novel framework 
to adapt existing models with the effective propagation structure for detecting rumors from different domains and/or languages. 

To facilitate related fact-checking tasks in low-resource settings, domain adaption techniques are utilized to detect fake news~\cite{wang2018eann, yuan2021improving, zhang2020bdann, silva2021embracing} by learning features from multi-modal data such as texts and images. 
\cite{lee2021towards} proposed a simple way of leveraging the perplexity score obtained from pre-trained language models (PLMs) for the few-shot fact-checking task. 
Different from these works of adaption on multi-modal data and transfer learning of PLMs, we focus on language and domain adaptation to detect rumors from low-resource microblog posts corresponding to breaking events. 

Contrastive learning (CL) aims to enhance representation learning by maximizing the agreement among the same types of instances and distinguishing from the others with different types~\cite{wang2020understanding}. 
In recent years, CL has achieved great success in unsupervised visual representation learning~\cite{chen2020simple, he2020momentum, chen2020big}
Besides computer vision, recent studies suggest that CL is promising in the semantic textual similarity~\cite{gao2021simcse, yan2021consert}, stance detection~\cite{mohtarami2019contrastive}, short text clustering~\cite{zhang2021supporting}, unknown intent detection~\cite{ lin2021boosting}, and abstractive summarization~\cite{liu2021simcls}, etc. However, the above CL frameworks are specifically proposed to augment unstructured textual data such as sentences and documents, which are not suitable for the low-resource rumor detection task considering claims together with more complex propagation structures of community response. 

This is a significant extension of the first contrastive approach for low-resource rumor detection on social media using a structure-based framework in our recent work~\cite{lin2022detect}. Since the publication of the work, several similar studies~\cite{yue2022contrastive, ma2022curriculum} with domain-adaptive contrastive learning on the fact-checking discipline have been conducted that build upon our findings. However, these studies have just focused on aligning different domains, while not adequately considering the distinctive informative features that may exist between different target samples and exploring effective augmentation strategies for structured data in the target-wise contrastive paradigm.


{
\begin{figure}[t]
    \begin{center}
    \resizebox{0.25\textwidth}{!}{\includegraphics{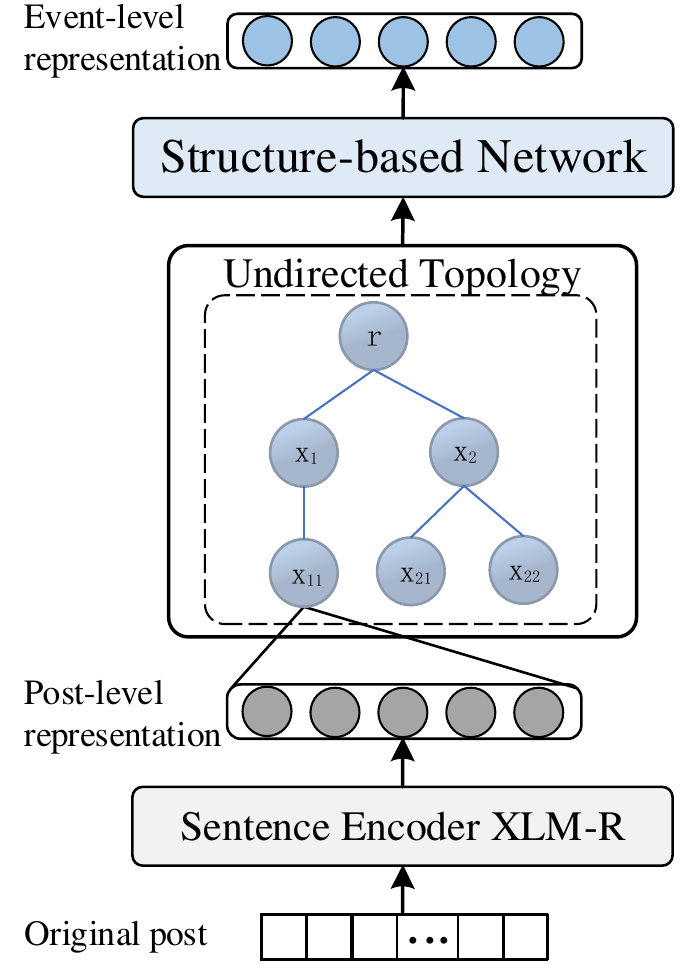}}
    \end{center}
    \caption{The overall architecture of our proposed method. For source and small target training data, we first obtain post-level representations after cross-lingual sentence encoding, then train the structure-based network (i.e., Multi-scale GCNs) with the unified contrastive objective. For target test data, we extract the event-level representations to detect rumors. }
\label{fig:method}    
\end{figure}}

\section{Problem Statement}
In this work, we define the low-resource rumor detection task as: Given a well-resourced dataset as source, classify each event in the target low-resource dataset as a rumor or not, where the source and target data are from different domains and/or languages. Specifically, we define a well-resourced source dataset for training as a set of events $\mathcal{D}_s = \{C_1^s, C_2^s, \cdots, C_{M}^s\}$, where $M$ is the number of source events. Each event $C^s=(y,c,\mathcal{T}(c))$ is a tuple representing a given claim $c$ which is associated with a veracity label $y \in \{\text{rumor}, \text{non-rumor}\}$, and ideally all its relevant responsive microblog post in chronological order, i.e., $\mathcal{T}(c) = \{c, x_{1}^s,x_{2}^s,\cdots,x_{|C|}^{s}\}$\footnote{$c$ is equivalent to $x_0^s$.}, where 
$|C|$ is the number of responsive tweets in the conversation thread. 
For the target dataset with low-resource domains and/or languages, we consider a much smaller dataset for training $\mathcal{D}_t = \{C_1^t, C_2^t, \cdots, C_{N}^t\}$, where $N (N \ll M)$ is the number of target events and each $C^t=(y,c',\mathcal{T}(c'))$ 
has a similar composition structure to the source dataset.

We formulate the task of low-resource rumor detection as a supervised classification problem that trains a domain/language-agnostic classifier $f(\cdot)$ adapting the features learned from source datasets to that of the target events, that is, $f(C^t| \mathcal{D}_s) \rightarrow y$. Note that although the tweets are notated sequentially, there are connections among them based on their responsive relationships~\cite{ma2018rumor}.

\section{Our Approach}
In this section, we introduce our unified contrastive transfer framework with propagation structure for adapting the features captured from the well-resourced data to detect rumors from low-resource events, which considers cross-lingual and cross-domain transfer. Figure~\ref{fig:method}--\ref{contrastive_learning} show an overview of our backbone model and training paradigm, which will be depicted in the following subsections.

{
\begin{figure}[t]
    \begin{center}
    \setlength{\belowcaptionskip}{0.1cm}
    \resizebox{0.49\textwidth}{!}{\includegraphics{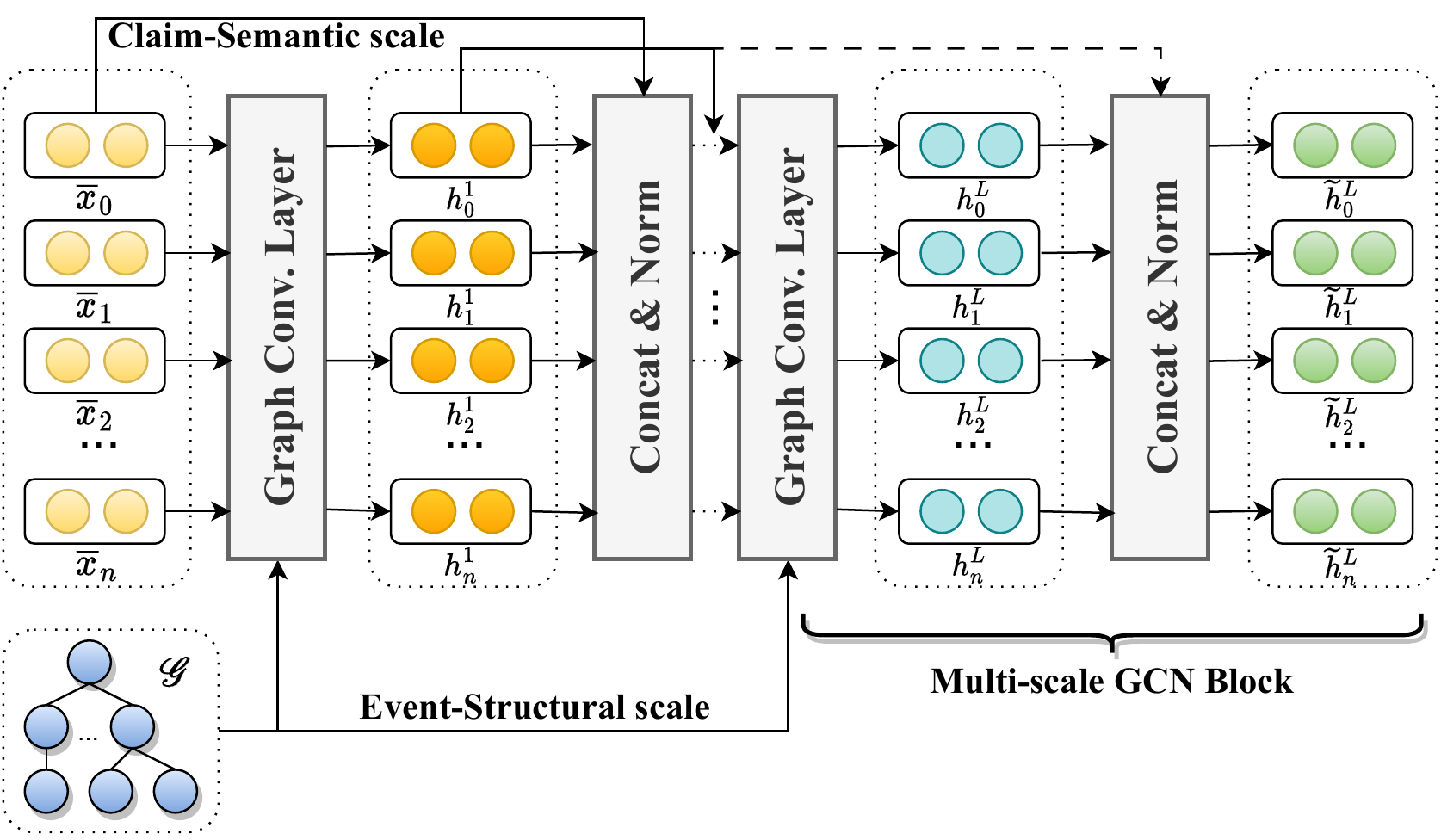}}
    \end{center}
    \caption{The Multi-scale GCNs for propagation structure representations.}
\label{multi-scale gcn}    
\end{figure}}

\subsection{Cross-lingual Sentence Encoder}
Given a post in an event that could be either from source or target data, to map it into a shared semantic space where the source and target languages are semantically aligned, we utilize XLM-RoBERTa~\cite{conneau2019unsupervised} (XLM-R) to model the context interactions among tokens in the sequence for the sentence-level representation:
{\setlength{\abovedisplayskip}{0.1cm}
\setlength{\belowdisplayskip}{0.1cm}
\begin{equation}\label{eq:1}
    \bar{x} = \textit{XLM-R}({x})
\end{equation}
}where ${x}$ is the original post, and we obtain the post-level representation $\bar{x}$ using the output state of the ${<s>}$ token in XLM-R. We thus denote the representation of posts in the source event $C^s$ and the target event $C^t$ as a matrix $X^s$ and $X^t$ respectively: $X^* = {[\bar{x}_0^*, \bar{x}_1^*,\bar{x}_2^*,...,\bar{x}_{|X^*|-1}^*]}^{\top}; *\in\{s,t\}$, where $X^s \in \mathbb{R}^{m\times d}$ and $X^t \in \mathbb{R}^{n \times d}$, $d$ is the dimension of the output state of the sentence encoder.

\subsection{Propagation Structure Representation}
\label{GCN}
On top of the sentence encoder, different from the directed tree structure modeling in previous work~\cite{ma2018rumor, bian2020rumor}, we first represent the propagation of each claim as an undirected propagation topology to explore the full-duplex interaction patterns between responsive
nodes with the expressive capacity of graph neural networks~\cite{kipf2016semi}. To fully utilize the claim's abundant information while preventing off-topic coherence that strays from the claim's main point in the propagation structure, as illustrated in Figure~\ref{multi-scale gcn}, we exploit a simple but effective Multi-scale Graph Convolutional Network to integrate both the claim semantics and the social context information for the subsequent contrastive training paradigm.

Given an event and its initialized embedding matrix $C^*, X^*; * \in \{s,t\}$, We model the propagation thread of the event as an undirected graph topology $\mathcal{G}=\langle V,E \rangle$, where $V$ consists of the event claim and all its relevant responsive posts as nodes and $E$ refers to a set of undirected edges corresponding to the response relation among the nodes in $V$. For example, for any ${x}_i,{x}_j \in {V}$, ${x}_i \to {x}_j$ and ${x}_j \to {x}_i$ exist if they have responsive relationships.

We transform the edge $E$ into a symmetric adjacency matrix $\textbf{A} \in {\{0,1\}}^{|{V}| \times |{V}|}$, where $\textbf{A}_{i,j} = 1$ if $\textbf{x}_{i}$ has a responsive relationship with $\textbf{x}_{j}$ or $i=j$, else $\textbf{A}_{i,j} = 0$. Then we utilize a layer-wise propagation rule to update the node vector at the $l^{\text{th}}$ layer: 
{\setlength{\abovedisplayskip}{0.1cm}
\setlength{\belowdisplayskip}{0.1cm}
\begin{equation}
    H^{(l+1)} = \operatorname{ReLU}\left(\hat{\textbf{A}} \cdot \tilde{H}^{(l)} \cdot W^{(l)}\right)
    \label{equ:GCN}
\end{equation}} where $\hat{\textbf{A}} = \textbf{D}^{-1/2} \textbf{A} \textbf{D}^{-1/2}$ is the symmetric normalized adjacency matrix, $\textbf{D}$ denotes the degree matrix of $\textbf{A}$. $W^{(l)} \in \mathbb{R}^{d^{(l)} \times d^{(l+1)}}$ is
a layer-specific trainable transformation matrix. In terms of $\tilde{H}^{(l)} = {[\tilde{h}_0^{(l)}, \tilde{h}_1^{(l)},\tilde{h}_2^{(l)},...,\tilde{h}_{|X^*|-1}^{(l)}]}^{\top}$, we employ a residual connection~\cite{he2016deep} around each graph convolutional layers for the multi-scale information fusion from both the claim-semantic scale and the event-structural scale, to obtain the refined representations:
\begin{equation}
    \tilde{H}^{(l)} = \operatorname{LayerNorm}\left(H^{(l)} || h_{0}^{(l-1)}\right)
\end{equation} where $h_{0}^{(l-1)} \in \mathbb{R}^{ d^{(l-1)}}$ is the hidden representations of the claim at the $(l-1)^{\text{th}}$ layer, and $||$ is the concatenation operation with the broadcast mechanism.
$H^{(0)}$ and $\tilde{H}^{(0)}$ are both initialized as $X^*$. 

For the Multi-scale GCN model with $L$-layers, we obtain the final node representation $\tilde{H}^{(L)}$ and jointly capture the opinions expressed in the propagation thread via mean-pooling:
{
\begin{equation}\label{eq:gcn}
    o = \operatorname{mean-pooling}(\tilde{H}^{(L)})
\end{equation}} where $o \in \mathbb{R}^{d^{(L)}}$ is the event-level structural representation of the entire propagation thread, $d^{(L)}$ is the output dimension of GCN.

{
\begin{figure}[t]
    \begin{center}
    \setlength{\belowcaptionskip}{0.1cm}
    \resizebox{0.49\textwidth}{!}{\includegraphics{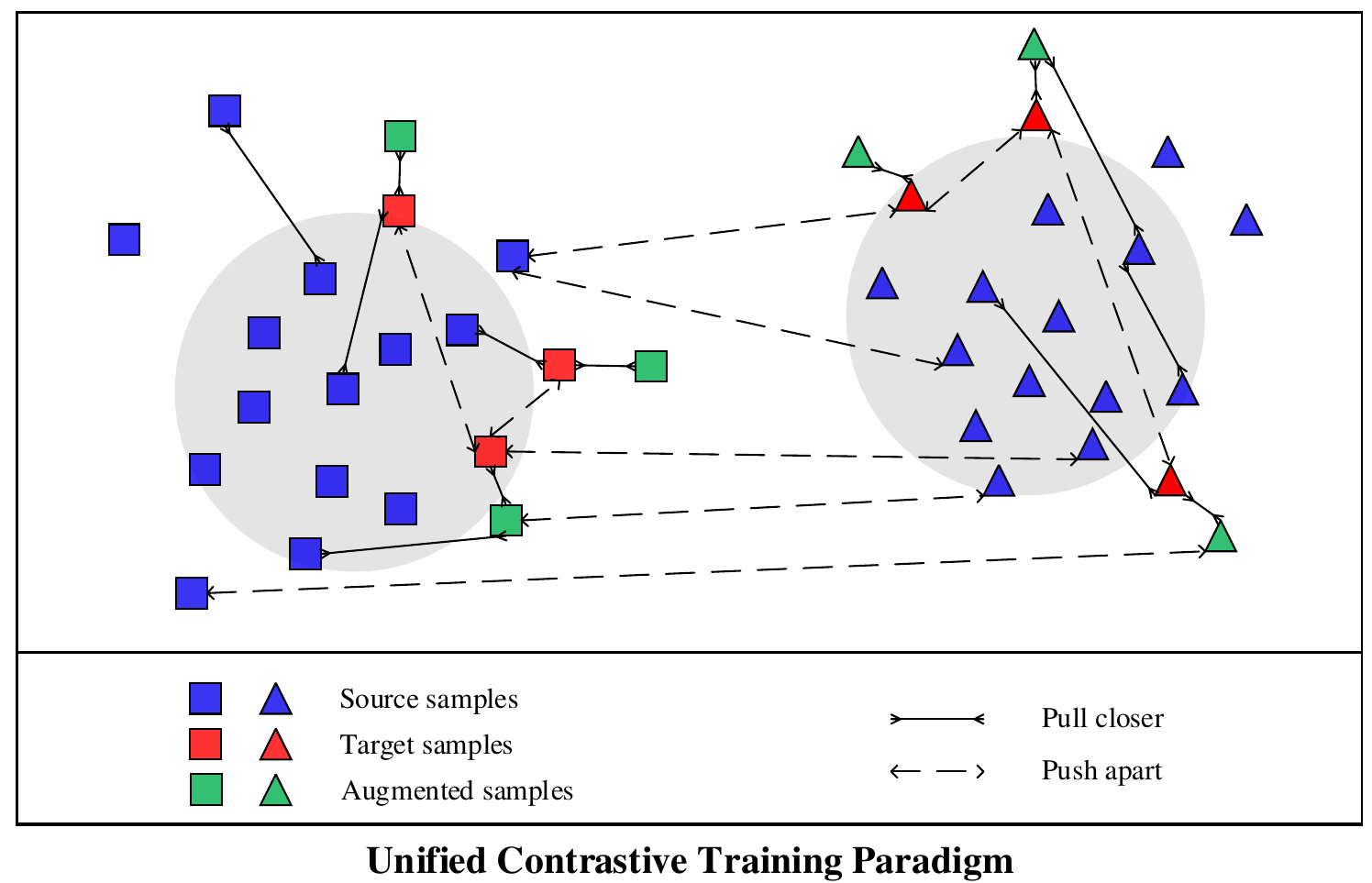}}
    \end{center}
    \caption{The unified contrastive training paradigm: domain-adaptive and target-wise contrastive learning.}
\label{contrastive_learning}    
\end{figure}}

\subsection{Domain-Adaptive Contrastive Training}
To align the representation space of rumor-indicative signals from different domains and languages, we present a novel training paradigm to exploit the labeled data including rich sourced data and small-scaled target data to adapt our model on target domains and languages. The core idea is to make the representations of source and target events from the same class closer while keeping representations from different classes far away, as shown in Figure~\ref{contrastive_learning}.

Given an event $C_i^s$ from the source data, we firstly obtain the language-agnostic encoding for all the involved posts (see Eq.~\ref{eq:1}) as well as 
the propagation structure representation $o_i^s$ (see Eq.~\ref{eq:gcn}) which is then fed into a \textit{softmax} function to make rumor predictions. 
Then, we learn to minimize the cross-entropy loss between the prediction and the ground-truth label $y_i^s$: 
{\setlength{\abovedisplayskip}{0.1cm}
\setlength{\belowdisplayskip}{0.1cm}
\begin{equation}
    \mathcal{L}_{CE}^s = -\frac{1}{N^s} \sum\limits_{i=1}^{N^s} log(p_i)
    \label{equ:ce}
\end{equation}} where $N^s$ is the total number of source examples in the batch, $p_i$ is the probability of correct prediction. 
To make rumor representation in the source events more dicriminative, we propose a supervised contrastive learning objective to cluster the same class and separate different classes of samples:
{
\begin{equation}
\begin{split}
    \mathcal{L}_{SCL}^s=-\frac{1}{N^s}\sum\limits_{i=1}^{N^s} \frac{1}{N_{y_i^s}-1} \sum\limits_{j=1}^{N^s}\mathbbm{1}_{[i \ne j]}\mathbbm{1}_{[y_i^s = y_j^s]}\\
    log\frac{\emph{exp}(\text{sim}(o_i^s, o_j^s))}{\sum\limits_{k=1}^{N^s} \mathbbm{1}_{[i \ne k]} \emph{exp}(\text{sim}(o_i^s, o_k^s))}
\end{split}
\end{equation}} where $N_{y_i^s}$ is the number of source examples with the same label $y_i^s$ in the event $C_i^s$, and $\mathbbm{1}$ is the indicator. $\text{sim}(\cdot)$ is the normalized temperature-scaled cosine similarity function. 

For an event $C_i^t$ from the target data, we also compute the classification loss $\mathcal{L}_{CE}^t$ in the same manner as Eq.~\ref{equ:ce}. Although we projected the source and target languages into the same semantic space after sentence encoding, rumor detection not only relies on post-level features, but also on event-level contextual features. Without constraints, the structure-based network can only extract event-level features for all samples based on their final classification signals while these features may not be critical to the target domain and language. We make full use of the minor labels in the low-resource rumor data by parameterizing our model according to the contrastive objective between the source and target instances in the event-level representation space:
{
\begin{equation}
\begin{split}
  \mathcal{L}_{SCL}^t = -\frac{1}{N^t}\sum\limits_{i=1}^{N^t} \frac{1}{N_{y_i^t}} \sum\limits_{j=1}^{N^s}\mathbbm{1}_{[y_i^t = y_j^s]}\\
    log\frac{\emph{exp}(\text{sim}(o_i^t, o_j^s))}{\sum\limits_{k=1}^{N^s} \emph{exp}(\text{sim}(o_i^t, o_k^s))}
 \end{split}
\end{equation}} where $N^t$ is the total number of target examples in the batch and $N_{y_i^t}$ is the number of source examples with the same label $y_i^t$ in the event $C_i^t$. 
As a result, we project the source and target samples belonging to the same class closer than that of different categories, 
for feature alignment with minor annotation at the target domain and language. 

\subsection{Target-wise Contrastive Training}
A critical defect of the domain-adaptive contrastive training is that almost all the target instances with the same veracity are mapped into a small projection space by simply maximizing inter-class variance and minimizing intra-class variance. Therefore, when there is a limited amount of labeled target samples, the structure-derived event-level representations could be somehow collapsed~\cite{chen2021exploring} and less discriminative to identify individual target samples.
For instance, in the context of the COVID-19 domain, the learned rumor-indicative features for target low-resource data could just converge to the general patterns like ``not true" or ``joke" in well-resourced data of open domains, which may fail to detect unlabeled COVID-19 data with unseen domain-specific patterns like ``just flu" or ``it's bioweapon" - Denial opinions towards rumors that minimize the severity of COVID-19 or be fueled by conspiracy theories.
Thus it's important to make the target representation evenly distributed and discriminative with each other for more representative feature learning.
To this end, we further exploit a target-wise contrastive learning to distinguish individual target on top of the alignment between group of samples with different veracity classes, which reinforces it to preserve maximal rumor-indicative information of the target events.

Given the event-level structural representation of a target sample $o_i^t$, we perform the target-wise contrastive objective based on the augmented event-level target data. As $N^t$ events are randomly selected from $D_t$ during each training stage to create a mini-batch, we first augment each target event to construct a pair of positive samples, leading to $2N^t$ event-level representations.
To differentiate from other target samples, each target data sample
is taught to identify its corresponding augmented sample from a batch of $2(N^t-1)$ negative samples:
{
\begin{gather}
  \mathcal{L}_{TCL}^t = -\frac{1}{N^t}\sum\limits_{i=1}^{N^t}log \nonumber
    \\ \frac{\emph{exp}(\text{sim}(o_i^t, \overline{o}_i^t))}{\sum\limits_{k=1}^{N^t} \mathbbm{1}_{[i \ne k]} \left(\emph{exp}(\text{sim}(o_i^t, o_k^t))+\emph{exp}(\text{sim}(o_i^t, \overline{o}_k^t))\right)}
\end{gather}} where $\overline{o}_i^t$ denotes the augmented event-level target representation of ${o}_i^t$, which is generated with the data augmentation strategies that would be depicted in the following subsection. In summary, as shown in Figure~\ref{contrastive_learning}, the target-wise contrastive objective focuses on distinguishing different target events for uniformly distributed event-level representations, and meanwhile the domain-adaptive contrastive objective identifies distinct rumor veracity from different domains and/or languages. As a result, the representations can be further enhanced by capturing more target-specific informative signals and well-generalized on diverse low-resource breaking events.

\begin{figure}[t]
    \begin{center}
    \setlength{\belowcaptionskip}{0.1cm}
    \resizebox{0.49\textwidth}{!}{\includegraphics{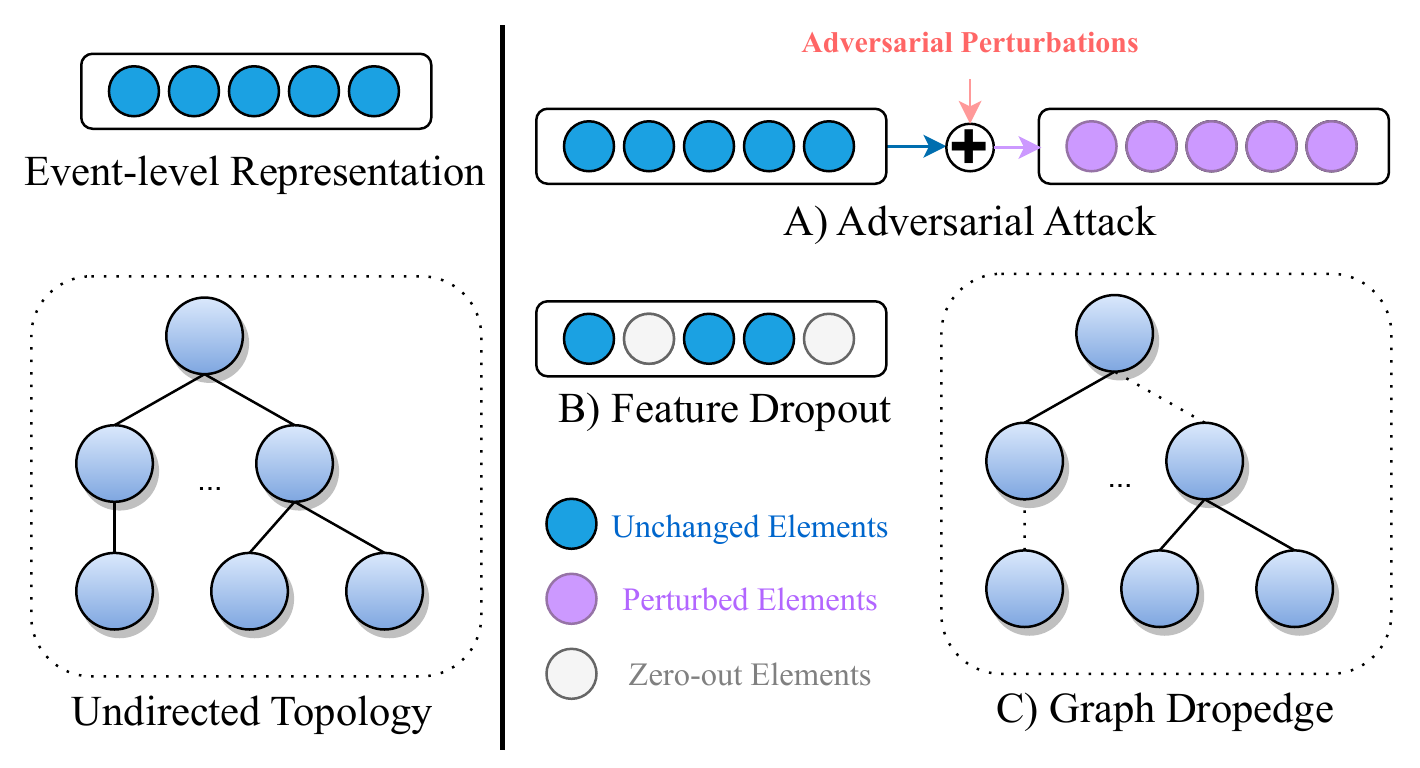}}
    \end{center}
    \caption{The three data augmentation strategies exploited in our framework: Adversarial Attack, Feature Dropout, and Graph Dropedge.}
\label{data_augment}    
\end{figure}

\subsection{Data Augmentation Strategies}
Data augmentation techniques were successfully utilized to enhance contrastive learning models~\cite{chen2020simple}, which involve creating different views or perspectives of the same data to be used as positive pairs in the target-wise contrastive learning process. In this study, we consider data augmentation from two perspectives: the encoding of the event-level representations and the modeling of the propagation structure. Thus we investigate three data augmentation techniques as shown in Figure~\ref{data_augment}. The first two strategies (i.e., Adversarial Attack~\cite{goodfellow2014generative, kurakin2016adversarial} and Feature Dropout~\cite{hinton2012improving}) are utilized to encode the event-level representations in our framework, which are widely utilized in recent studies~\cite{gao2021simcse,yan2021consert}.  
To model the inherent complexity and dynamic nature of rumor dissemination~\cite{rong2019dropedge}, we attempt to augment data based on the propagation structure of target events by masking some sampling edges in the undirected propagation structure as shown in the third strategy. 


\textbf{Adversarial Attack} Adversarial training is commonly used to improve the robustness of a model. To create an adversarial example, we apply Fast Gradient Value~\cite{rozsa2016adversarial} to approximate a worst-case perturbation at the event-level representations, where the gradient is normalized to represent the direction that significantly decreases the model’s prediction performance. Then we obtain the pseudo adversarial sample by adding the perturbation to the event-level representations.

\textbf{Feature Dropout} Dropout is a widely used regularization method that avoids overfitting. However, in this work, we also show its effectiveness as an augmentation strategy of event-level representations for contrastive learning. For this setting, we randomly drop elements in the event-level representations by a specific probability and set their values to zero.

\textbf{Graph Dropedge} Different from Adversarial Attack and Feature Dropout directly applied to the encoding of event-level representations, we further explore an augmentation strategy based on the propagation structure. In particular, we randomly remove edges from the input undirected graphs throughout each training period to produce the deformed copies by a specific probability, which then be input into the structure-based network, i.e., Multi-scale GCNs, for the augmented event-level representations.

\begin{algorithm}[t]
\small
  \caption{\textbf{Unified Contrastive Learning}}
  \label{algorithm}
  \begin{algorithmic}[1]
    \Require
      A small set of events $C_i^t$ in the target domain and language; A set of events $C_i^s$ in the source domain and language.
    \Ensure
      Assign rumor labels $y$ to given unlabeled target data.
	\State \textbf{for} each mini-batch $N^t$ of the target events $C_i^t$ \textbf{do}:
    \State \quad \textbf{for} each mini-batch $N^s$ of the source events $C_i^s$ \textbf{do}:
    \State \quad \quad Pass $C_i^*$ to the sentence encoder and then structure-based network to obtain its event-level feature $o_i^*$, where $* \in \{s, t\}$. 
    \State \quad \quad Compute the classification loss $\mathcal{L}_{CE}^*$ for source and target data, respectively.
    \State \quad \quad Data augmentation for target data to compute the target-wise contrastive loss $\mathcal{L}_{TCL}^t$ and update $\mathcal{L}_{CE}^t$.
    \State \quad \quad Compute the domain-adaptive contrastive loss $\mathcal{L}_{SCL}^*$.
    \State \quad \quad Compute the joint loss $\mathcal{L}^*$ as Eq.~\ref{joint_loss}.
    \State \quad \quad Jointly optimize all parameters of the model using the average loss $\mathcal{L} = \operatorname{mean}({\mathcal{L}^s}+{\mathcal{L}^t})$.
  \end{algorithmic}
\end{algorithm}

\subsection{Model Training}
We jointly train the model with the cross-entropy and contrastive objectives for the source and target training data:
{
\begin{equation}
    \label{joint_loss}
    \mathcal{L}^*=(1-\alpha)\mathcal{L}_{CE}^* + \alpha \left( \mathcal{L}_{SCL}^* + \mathbbm{1}_{[* = t]} \mathcal{L}_{TCL}^*\right); * \in \{s, t\}
\end{equation}} 
where $\alpha$ is a trade-off parameter, which is set to 0.5 in our experiments. Algorithm~\ref{algorithm} presents the training process of our approach. The framework is alternately trained using stochastic gradient descent with mini-batches~\cite{bordes2009sgd}. For each mini-batch of target training data, we traverse the source data by repeating Step 3-8 in  Algorithm~\ref{algorithm}. Firstly, We encode post-level representations, obtain the structure-derived event-level representations and compute traditional classification losses for source and target training data, respectively. After that, the data augmentation is conducted on target training data for the computation of the target-wise contrastive loss. And then the domain-adaptive contrastive loss is computed. In terms of Step 7, note that the training objective for the target data considers the target-wise contrastive loss in addition to the supervised contrastive loss and classification loss.
We set the number $L$ of the graph convolutional layer as 2. Parameters are updated through back-propagation~\cite{collobert2011natural} with the Adam optimizer~\cite{loshchilov2018decoupled}. The learning rate is initialized as 0.0001, and the dropout rate is 0.2. Early stopping~\cite{yao2007early} is applied to avoid overfitting.

\begin{table}[t] \Huge
\centering
\caption{Statistics of Datasets.}
\label{tab:statistics}
\resizebox{0.475\textwidth}{!}{
\begin{tabular}{l||cc||cccc}
\hline
\multirow{2}{*}{Dataset} & \multicolumn{2}{c|}{Source} & \multicolumn{4}{c}{Target}                          \\ \cline{2-7} 
                         & TWITTER      & WEIBO        & EngCovid & ChiCovid & CanCovid &AraCovid     \\ \hline \hline
\# of events             & 1154         & 4649         & 400             & 399           & 1481       &218       \\ \hline
\# of tree nodes         & 60409        & 1956449      & 406185          & 26687         & 68490       &99786     \\ \hline
\# of non-rumors         & 579          & 2336         & 148             & 146           & 920        &78       \\ \hline
\# of rumors             & 575          & 2313         & 252             & 253           & 561        &140       \\ \hline
Avg. time/tree    & 389h    & 1007h   & 2497h      & 248h     & 668h    &2154h     \\ \hline
Avg. depth/tree          & 11.67        & 49.85        & 143.03          & 4.31          & 9.98       &35.54      \\ \hline
Language                 & English      & Chinese      & English         & Chinese       & Cantonese & Arabic \\ \hline
Domain                   & \multicolumn{2}{c|}{Open}         & \multicolumn{4}{c}{COVID-19}      \\ \hline
\end{tabular}}
\end{table}

\section{Experiments}

\begin{table*}[t] \large
\centering
\begin{center}
\caption{Rumor detection results on the target test datasets Chinese-COVID19 and English-COVID19.}
\label{tab:main_results_1}
\resizebox{0.75\textwidth}{!}{
\begin{tabular}{l||cc|cc||cccc}
\hline
Target (Source)        & \multicolumn{4}{c|}{Chinese-COVID19 (TWITTER)}                                      & \multicolumn{4}{c}{English-COVID19 (WEIBO)}                                                            \\ \hline
\multirow{2}{*}{Model} 
                       & \multirow{2}{*}{Acc.} & \multirow{2}{*}{Mac-$\emph{F}_1$ } & Rumor          & Non-rumor      & \multirow{2}{*}{Acc.} & \multicolumn{1}{c|}{\multirow{2}{*}{Mac-$\emph{F}_1$ }} & Rumor          & Non-rumor      \\ \cline{4-5} \cline{8-9} 
                       &                       &                         & $\emph{F}_1$              & $\emph{F}_1$              &                       & \multicolumn{1}{c|}{}                        & $\emph{F}_1$              & $\emph{F}_1$              \\ \hline \hline
CNN              & 0.445                 & 0.402                   & 0.476          & 0.328              & 0.498                 & \multicolumn{1}{c|}{0.389}                   & 0.528          & 0.249              \\
RNN          & 0.463                 & 0.414                   & 0.498              & 0.329          & 0.510                 & \multicolumn{1}{c|}{0.388}                   & 0.533              & 0.243          \\\hdashline
RvNN                   & 0.514                 & 0.482                   & 0.538          & 0.426          & 0.540                 & \multicolumn{1}{c|}{0.391}                   & 0.534          & 0.247          \\
PLAN                   & 0.532                 & 0.496                   & 0.578          & 0.414          & 0.573                 & \multicolumn{1}{c|}{0.423}                   & 0.549          & 0.298          \\
BiGCN                  & 0.569                 & 0.508                   & 0.586          & 0.429          & 0.616                 & \multicolumn{1}{c|}{0.415}                   & 0.577          & 0.252          \\ \hline
DANN-RvNN              & 0.583                 & 0.498                   & 0.591          & 0.405          & 0.577                 & \multicolumn{1}{c|}{0.482}                   & 0.648          & 0.317          \\
DANN-PLAN              & 0.601                 & 0.507                   & 0.606          & 0.409          & 0.593                 & \multicolumn{1}{c|}{0.471}                   & 0.574          & 0.369          \\
DANN-BiGCN             & 0.629                 & 0.561                   & 0.616          & 0.506          & 0.618                 & \multicolumn{1}{c|}{0.510}                   & 0.676          & 0.344          \\ \hline
UCLR-RvNN               & 0.801                 & 0.743                   & 0.844          & 0.642          & 0.676                 & \multicolumn{1}{c|}{0.618}                   & 0.740          & 0.496          \\
UCLR-PLAN               & 0.849                 & 0.816                   & 0.871          & 0.760          & 0.724                 & \multicolumn{1}{c|}{0.651}                   & 0.769          & 0.533          \\
UCLR-BiGCN              & {0.885}        & {0.867}          & {0.898} & {0.835} & {0.769}        & \multicolumn{1}{c|}{{0.687}}          & {0.815} & \textbf{0.559} \\ \hline
UCLR              & \textbf{0.895}        & \textbf{0.883}          & \textbf{0.916} & \textbf{0.851} & \textbf{0.773}        & \multicolumn{1}{c|}{\textbf{0.692}}          & \textbf{0.827} & {0.556} \\
\hline
\end{tabular}}
\end{center}
\end{table*}

\subsection{Datasets}
The focus of this work, as well as in many previous studies~\cite{ma2017detect, ma2018rumor, khoo2020interpretable, bian2020rumor}, is rumors on social media, instead of ``fake news" strictly defined as a news article published by a news outlet that is verifiably false~\cite{shu2017fake,zubiaga2018detection}.
To our knowledge, there are no public benchmarks available for detecting low-resource rumors with propagation tree structure in tweets. In this paper, we consider a breaking event COVID-19 as a low-resource domain and collect relevant rumors and non-rumors respectively from Twitter in English, Cantonese and Arabic, and Sina Weibo in Chinese. For the data from Twitter (English-COVID19, Cantonese-COVID19 and Arabic-COVID19), we resort two COVID-19 rumor datasets~\cite{alam2021fighting, ke2020novel} of tweets, which only contains textual claims without propagation threads. We extend each claim by collecting its propagation thread via Twitter academic API with a twarc2 package\footnote{\url{https://twarc-project.readthedocs.io/en/latest/twarc2_en_us/}} in python. For data from Sina Weibo (Chinese-COVID19), data annotation similar to \cite{ma2016detecting}, a set of rumorous claims is gathered from the Sina community management center\footnote{\url{https://service.account.weibo.com/}} and non-rumorous claims by randomly filtering out the posts that are not reported as rumors. Weibo API is utilized to collect all the repost/reply messages towards each claim. All the datasets contain two binary labels: Rumor and Non-rumor. The statistics of the six datasets are shown in Table~\ref{tab:statistics}, where EngCovid, ChiCovid, CanCovid and AraCovid denote the English-COVID19, Chinese-COVID19, Cantonese-COVID19 and Arabic-COVID19, respectively.

\subsection{Experimental Setup}
We compare our model and several state-of-the-art baseline methods described below. 1) \textbf{CNN}: A CNN-based model for misinformation identification~\cite{yu2017convolutional} by framing the relevant posts as a fixed-length sequence; 2) \textbf{RNN}: A RNN-based rumor detection model~\cite{ma2016detecting} with GRU for feature learning of relevant posts over time; 3) \textbf{RvNN}: A rumor detection approach based on tree-structured recursive neural networks \cite{ma2018rumor} that learn rumor representations guided by the propagation structure; 4) \textbf{PLAN}: A transformer-based model \cite{khoo2020interpretable} for rumor detection to capture long-distance interactions between any pair of involved tweets; 5) \textbf{BiGCN}: A GNN-based model~\cite{bian2020rumor} based on directed conversation trees to learn higher-level representations; 6) \textbf{DANN-*}: We employ and extend an existing few-shot learning technique, domain-adversarial neural network~\cite{ganin2016domain}, based on the structure-based model where * could be RvNN, PLAN, and BiGCN; 7) \textbf{UCLR-*}: our proposed unified contrastive learning objectives on top of RvNN, PLAN, or BiGCN; 8) \textbf{UCLR}: our proposed unified propagation-aware contrastive transfer framework with multi-scale GCNs.

As the key insight to fill the low-resource gap is to relieve the limitation imposed by the specific language resource dependency besides the specific domain,
in this work, we consider the most challenging case, i.e., detecting events (i.e., target) from a new domain and language. 
Specifically, we use TWITTER~\cite{ma2017detect} and WEIBO~\cite{ma2016detecting} datasets as the source data; Chinese-COVID19, English-COVID19, Cantonese-COVID19 and Arabic-COVID19 datasets as the target\footnote{English-COVID19 and Chinese-COVID19 datasets has already been public at \url{https://github.com/DanielLin97/ACLR4RUMOR-NAACL2022}}. 
We use accuracy and macro-averaged F1, as well as class-specific F1 scores as the evaluation metrics.

\begin{table*}[t]\large
\centering
\begin{center}
\caption{Rumor detection results on the target test datasets Cantonese-COVID19 and Arabic-COVID19. The symbol $\cdot \texttt{|} \cdot$ for the transfer models denotes the different performance from the models trained on different source datasets, TWITTER and WEIBO, respectively.}
\label{tab:main_results_2}
\resizebox{1.\textwidth}{!}{
\begin{tabular}{l||cccc||cccc}
\hline
Target                 & \multicolumn{4}{c|}{Cantonese-COVID19}                                                              & \multicolumn{4}{c}{Arabic-COVID19}                                                                                    \\ \hline
\multirow{2}{*}{Model} & \multirow{2}{*}{Acc.} & \multicolumn{1}{c|}{\multirow{2}{*}{Mac-$\emph{F}_1$}} & Rumor       & Non-rumor   & \multirow{2}{*}{Acc.} & \multicolumn{1}{c|}{\multirow{2}{*}{Mac-$\emph{F}_1$}} & Rumor                            & Non-rumor   \\ \cline{4-5} \cline{8-9} 
                       &                       & \multicolumn{1}{c|}{}                        & $\emph{F}_1$          & $\emph{F}_1$          &                       & \multicolumn{1}{c|}{}                        & $\emph{F}_1$                               & $\emph{F}_1$          \\ \hline \hline
CNN                    & 0.508                 & \multicolumn{1}{c|}{0.347}                   & 0.272       & 0.422       & 0.556                 & \multicolumn{1}{c|}{0.430}                   & \multicolumn{1}{c|}{0.632}       & 0.227       \\
RNN                    & 0.488                 & \multicolumn{1}{c|}{0.371}                   & 0.341       & 0.401       & 0.560                 & \multicolumn{1}{c|}{0.463}                   & \multicolumn{1}{c|}{0.687}       & 0.238       \\ \hdashline
RvNN                   & 0.535                 & \multicolumn{1}{c|}{0.451}                   & 0.334       & 0.568       & 0.565                 & \multicolumn{1}{c|}{0.467}                   & \multicolumn{1}{c|}{0.694}       & 0.239       \\
PLAN                   & 0.544                 & \multicolumn{1}{c|}{0.459}                   & 0.289       & 0.629       & 0.573                 & \multicolumn{1}{c|}{0.470}                   & \multicolumn{1}{c|}{0.641}       & 0.298       \\
BiGCN                  & 0.538                 & \multicolumn{1}{c|}{0.504}                   & 0.383       & 0.625       & 0.586                 & \multicolumn{1}{c|}{0.487}                   & \multicolumn{1}{c|}{0.698}       & 0.276       \\ \hline
DANN-RvNN              & 0.499\texttt{|}0.564           & \multicolumn{1}{c|}{0.465\texttt{|}0.539}             & 0.359\texttt{|}0.437 & 0.570\texttt{|}0.641 & 0.612\texttt{|}0.612           & \multicolumn{1}{c|}{0.547\texttt{|}0.547}             & \multicolumn{1}{c|}{0.713\texttt{|}0.713} & 0.381\texttt{|}0.381 \\
DANN-PLAN              & 0.531\texttt{|}0.572           & \multicolumn{1}{c|}{0.473\texttt{|}0.522}             & 0.339\texttt{|}0.370 & 0.607\texttt{|}0.673 & 0.636\texttt{|}0.631           & \multicolumn{1}{c|}{0.568\texttt{|}0.555}             & \multicolumn{1}{c|}{0.717\texttt{|}0.736} & 0.419\texttt{|}0.374 \\
DANN-BiGCN             & 0.591\texttt{|}0.631           & \multicolumn{1}{c|}{0.575\texttt{|}0.587}             & 0.539\texttt{|}0.454 & 0.611\texttt{|}0.720 & 0.642\texttt{|}0.665           & \multicolumn{1}{c|}{0.563\texttt{|}0.563}             & \multicolumn{1}{c|}{0.744\texttt{|}0.773} & 0.381\texttt{|}0.353 \\ \hline
UCLR-RvNN              & 0.571\texttt{|}0.670           & \multicolumn{1}{c|}{0.501\texttt{|}0.627}             & 0.323\texttt{|}0.499 & 0.679\texttt{|}0.754 & 0.659\texttt{|}0.678           & \multicolumn{1}{c|}{0.611\texttt{|}0.636}             & \multicolumn{1}{c|}{0.732\texttt{|}0.756} & 0.489\texttt{|}0.516 \\
UCLR-PLAN              & 0.650\texttt{|}0.703           & \multicolumn{1}{c|}{0.652\texttt{|}0.644}             & 0.599\texttt{|}0.509 & 0.704\texttt{|}0.779 & 0.686\texttt{|}0.690           & \multicolumn{1}{c|}{0.587\texttt{|}0.643}             & \multicolumn{1}{c|}{0.780\texttt{|}0.773} & 0.393\texttt{|}0.512 \\
UCLR-BiGCN             & 0.685\texttt{|}0.713           & \multicolumn{1}{c|}{0.656\texttt{|}0.692}             & 0.569\texttt{|}0.631 & 0.742\texttt{|}0.752 & 0.673\texttt{|}0.714           & \multicolumn{1}{c|}{0.618\texttt{|}0.665}             & \multicolumn{1}{c|}{0.759\texttt{|}0.782} & 0.477\texttt{|}0.548 \\ \hline
UCLR                   & 0.730\texttt{|}\textbf{0.733}           & \multicolumn{1}{c|}{\textbf{0.705}\texttt{|}0.701}             & \textbf{0.632}\texttt{|}0.612 & 0.777\texttt{|}\textbf{0.789} & 0.713\texttt{|}\textbf{0.732}           & \multicolumn{1}{c|}{0.670\texttt{|}\textbf{0.687}}             & \multicolumn{1}{c|}{0.786\texttt{|}\textbf{0.797}} & 0.554\texttt{|}\textbf{0.577} \\ \hline
\end{tabular}}
\end{center}
\end{table*}

\subsection{Implementation Details}
We run all of our experiments on one single NVIDIA Tesla V100 GPU. We set the total batch size to 64, where the batch size of source samples is set to 32, the same as target samples. The hidden and output dimensions of each node in the structure-based network are set to 512 and 128, respectively. Since the focus in this paper is primarily on better leveraging the contrastive learning for domain and language adaptation on top of event-level representations, we choose the off-the-shelf multilingual PLM $\text{XLM-R}_{\textit{Base}}$ (Layer number = 12, Hidden dimension = 768, Attention head = 12, 270M params) as our sentence encoder for language-agnostic representations at the post level. To conduct five-fold cross-validation on the target dataset in our low-resource settings, we use each fold of the target dataset in turn for training with the whole source dataset, and test on the rest of the target dataset. The average runtime for our approach on five-fold cross-validation in one iteration is about 3 hours. The number of total trainable parameters is 562,818 for our model. We implement our model with pytorch\footnote{\url{pytorch.org}}.

\subsection{Rumor Detection Performance}
Table \ref{tab:main_results_1} shows the performance of our proposed method versus all the compared methods on the Chinese-COVID19 and English-COVID19 test sets, respectively. And Table~\ref{tab:main_results_2} further demonstrates the performance of all the compared models on the Cantonese-COVID19 and Arabic-COVID19 datasets. It is observed that the performances of the baselines in the first group are obviously poor due to ignoring intrinsic structural patterns. To make fair comparisons, all baselines are employed with the same cross-lingual sentence encoder of our framework as inputs. Other state-of-the-art baselines exploit the structural property of community wisdom on social media, which confirms the necessity of propagation structure representations aware in our framework. 

Among the structure-based baselines in the second group, due to the representation power of message-passing architectures and tree structures, PLAN and BiGCN outperform RvNN with only limited labeled target data for training. The third group shows the results for DANN-based methods with pre-determined training datasets TWITTER and WEIBO. It improves the performance of structure-based baselines in general since it extracts cross-domain features between source and target datasets via generative adversarial nets~\cite{goodfellow2014generative}. 

In contrast, our proposed UCLR-based framework on top of existing structure-based approaches in the fourth group achieves superior performance among all their counterparts ranging from 24.5\% (13.6\%) to 30.9\% (18.0\%) in terms of Macro F1 score on Chinese-COVID19 (English-COVID19) datasets in Table~\ref{tab:main_results_1}, and the similar phenomenon could be observed in Table~\ref{tab:main_results_2}, which suggests their strong judgment on low-resource rumors from different domains/languages. And the choice of propagation structure representation is orthogonal to our proposed framework that can be easily replaced with any existing structure-based models without any other change to our unified contrastive learning architecture. Meanwhile, it can be seen from Table~\ref{tab:main_results_2} that, for the same target data, our framework performs generally better when utilizing WEIBO as the source data. The plausible reason might be that WEIBO has a relatively larger amount of training data than TWITTER so our domain-adaptive contrastive learning could make full use of the well-resourced data for few-shot transfer.  

Our perfect model UCLR performs the best among all the baselines, even much better than the three UCLR-based variants, by mining effective clues simultaneously from the post semantics and the structural property via multi-scale encoding for conversation threads. Furthermore, the structure-based counterparts generally have more parameters and complex structures (UCLR-BiGCN with total trainable parameters 1,117,954) than Multi-scale GCNs of UCLR framework with total trainable parameters 562,818. Although such complex structure-based networks like BiGCN may show promising performance on the mono-domain and mono-lingual training corpora, their generalization ability in cross-domain and cross-lingual settings may be compromised. This is because excessively complex models may overfit the training set data, leading to inaccurate generalization to new target data. This also justifies the complementary of our proposed Multi-scale GCNs backbone and the UCLR training paradigm. In summary, the main results indicate that the unified contrastive learning framework can effectively transfer knowledge from the source to target data at the event level, and substantiate our method is model-agnostic for different structure-based networks. For a more clear qualitative analysis of the effectiveness of the domain-adaptive contrastive learning and the target-wise contrastive learning, we further provide the ablative test on the unified contrastive transfer framework UCLR in the following subsection Sec.~\ref{ablation}.

\subsection{Ablative study}\label{ablation}
We perform ablation studies based on our proposed approach UCLR, where the performance from models trained with TWITTER as the source data is shown in Table~\ref{ablative_1} and that with WEIBO as the source data is shown in Table~\ref{ablative_2}. For the cross-domain and cross-lingual settings, we use Chinese-COVID19, Cantonese-COVID19, and Arabic-COVID19 as the target data when TWITTER is utilized as the source data, and we use English-COVID19, Cantonese-COVID19, and Arabic-COVID19 as the target data when WEIBO is utilized as the source data.

\begin{table}[]\large
\caption{RESULTS OF THE ABLATION STUDY OF UCLR with TWITTER as the source data.}
\label{ablative_1}
\resizebox{0.48\textwidth}{!}{\begin{tabular}{l||cc|cc|cc}
\hline
Source     & \multicolumn{6}{c}{TWITTER}                                                                                        \\ \hline
Target     & \multicolumn{2}{c|}{ChiCovid} & \multicolumn{2}{c|}{CanCovid} & \multicolumn{2}{c}{AraCovid} \\ \hline
Model      & Acc.   & \multicolumn{1}{c|}{Mac-$\emph{F}_1$} & Acc.    & \multicolumn{1}{c|}{Mac-$\emph{F}_1$}  & Acc.            & Mac-$\emph{F}_1$           \\ \hline \hline
BiGCN(T)   & 0.569  & \multicolumn{1}{c|}{0.508}  & 0.538   & \multicolumn{1}{c|}{0.504}   & 0.586           & 0.487            \\
BiGCN(S)   & 0.578  & \multicolumn{1}{c|}{0.463}  & 0.562   & \multicolumn{1}{c|}{0.541}   & 0.631           & 0.536            \\
BiGCN(S,T) & 0.693  & \multicolumn{1}{c|}{0.472}  & 0.576   & \multicolumn{1}{c|}{0.558}   & 0.655           & 0.539            \\ \hline
DANN-BiGCN & 0.629  & \multicolumn{1}{c|}{0.561}  & 0.591   & \multicolumn{1}{c|}{0.575}   & 0.642           & 0.563            \\
ACLR-BiGCN & 0.873  & \multicolumn{1}{c|}{0.861}  & 0.653   & \multicolumn{1}{c|}{0.617}   & 0.671           & 0.579            \\
UCLR-BiGCN & 0.885  & \multicolumn{1}{c|}{0.867}  & 0.685   & \multicolumn{1}{c|}{0.656}   & 0.673           & 0.618            \\ \hline
$\text{UCLR}_{\text{Adv}}$  & 0.890  & \multicolumn{1}{c|}{0.871}  & 0.718   & \multicolumn{1}{c|}{0.697}   & 0.709           & 0.616            \\
$\text{UCLR}_{\text{Dropout}}$   & 0.888  & \multicolumn{1}{c|}{0.869}  & 0.721   & \multicolumn{1}{c|}{0.702}   & 0.699           & 0.580            \\
$\text{UCLR}_{\text{DropEdge}}$   & 0.895  & \multicolumn{1}{c|}{0.883}  & 0.730   & \multicolumn{1}{c|}{0.705}   & 0.713           & 0.670            \\ \hline
\end{tabular}}

\end{table}

\begin{table}[]\large
\caption{RESULTS OF THE ABLATION STUDY OF UCLR with WEIBO as the source data.}
\label{ablative_2}
\resizebox{0.48\textwidth}{!}{\begin{tabular}{l||cc|cc|cc}
\hline
Source     & \multicolumn{6}{c}{WEIBO}                                                                                          \\ \hline
Target     & \multicolumn{2}{c|}{ChiCovid} & \multicolumn{2}{c|}{CanCovid} & \multicolumn{2}{c}{AraCovid} \\ \hline
Model      & Acc.   & \multicolumn{1}{c|}{Mac-$\emph{F}_1$} & Acc.    & \multicolumn{1}{c|}{Mac-$\emph{F}_1$}  & Acc.            & Mac-$\emph{F}_1$           \\ \hline \hline
BiGCN(T)   & 0.616  & \multicolumn{1}{c|}{0.415}  & 0.538   & \multicolumn{1}{c|}{0.504}   & 0.586           & 0.487            \\
BiGCN(S)   & 0.578  & \multicolumn{1}{c|}{0.463}  &  0.562       & \multicolumn{1}{c|}{0.525}        & 0.612           & 0.506            \\
BiGCN(S,T) & 0.693  & \multicolumn{1}{c|}{0.472}  &  0.581       & \multicolumn{1}{c|}{0.538}        & 0.633           & 0.519            \\ \hline
DANN-BiGCN & 0.618  & \multicolumn{1}{c|}{0.510}  & 0.631   & \multicolumn{1}{c|}{0.587}   & 0.665           & 0.563            \\
ACLR-BiGCN & 0.765  & \multicolumn{1}{c|}{0.686}  & 0.698   & \multicolumn{1}{c|}{0.679}   & 0.707           & 0.624            \\
UCLR-BiGCN & 0.769  & \multicolumn{1}{c|}{0.687}  & 0.713   & \multicolumn{1}{c|}{0.692}   & 0.714           & 0.665            \\ \hline
$\text{UCLR}_{\text{Adv}}$  & 0.768  & \multicolumn{1}{c|}{0.691}  &  0.729       & \multicolumn{1}{c|}{0.697}        & 0.710           & 0.658            \\
$\text{UCLR}_{\text{Dropout}}$   & 0.771  & \multicolumn{1}{c|}{0.689}  &  0.727       & \multicolumn{1}{c|}{0.683}        & 0.718           & 0.676            \\
$\text{UCLR}_{\text{DropEdge}}$   & 0.773  & \multicolumn{1}{c|}{0.692}  & 0.733   & \multicolumn{1}{c|}{0.701}   & 0.732           & 0.687            \\ \hline
\end{tabular}}

\end{table}

{
\begin{figure*}[ht]
\centering
\subfigure[English-COVID19 (elapsed time)]{
\begin{minipage}[t]{0.33\linewidth}
\centering
\scalebox{0.85}{\includegraphics[width=6cm]{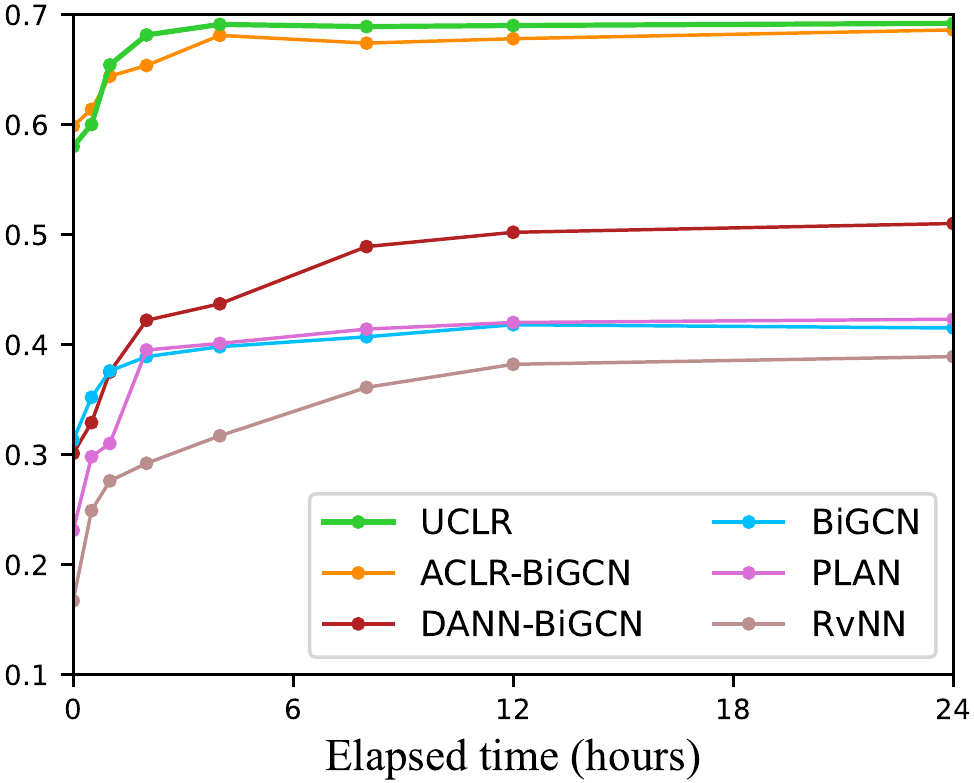}}
\label{fig:early_a}
\end{minipage}%
}%
\subfigure[Cantonese-COVID19 (elapsed time)]{
\begin{minipage}[t]{0.33\linewidth}
\centering
\scalebox{0.85}{\includegraphics[width=6cm]{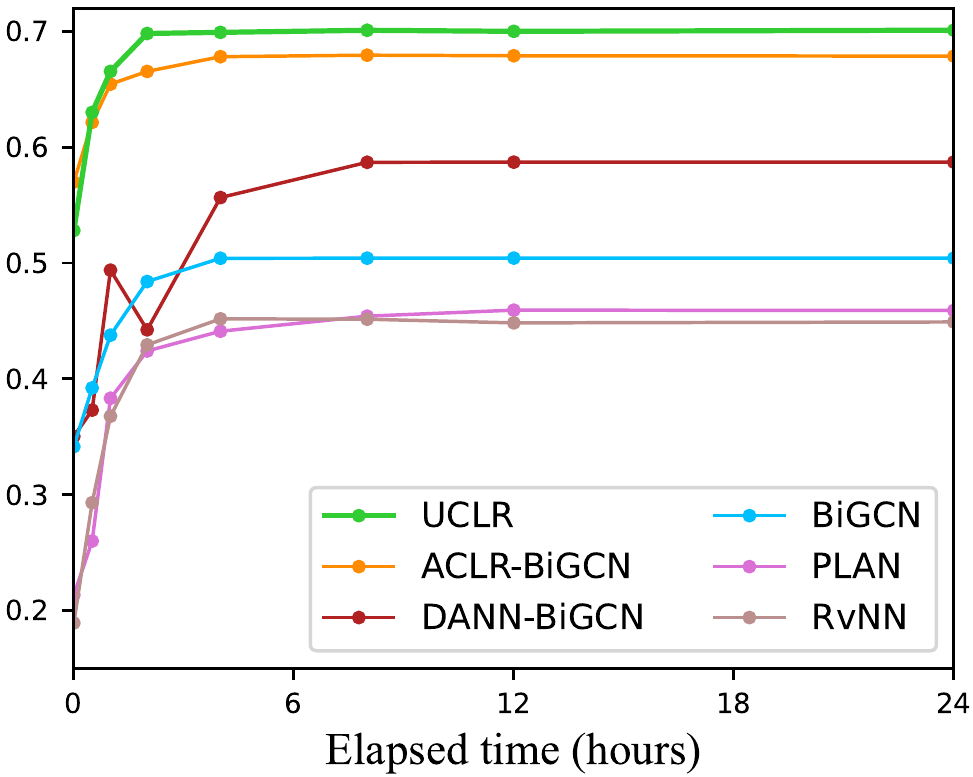}}
\label{fig:early_b}
\end{minipage}%
}%
\subfigure[Arabic-COVID19 (elapsed time)]{
\begin{minipage}[t]{0.33\linewidth}
\centering
\scalebox{0.85}{\includegraphics[width=6cm]{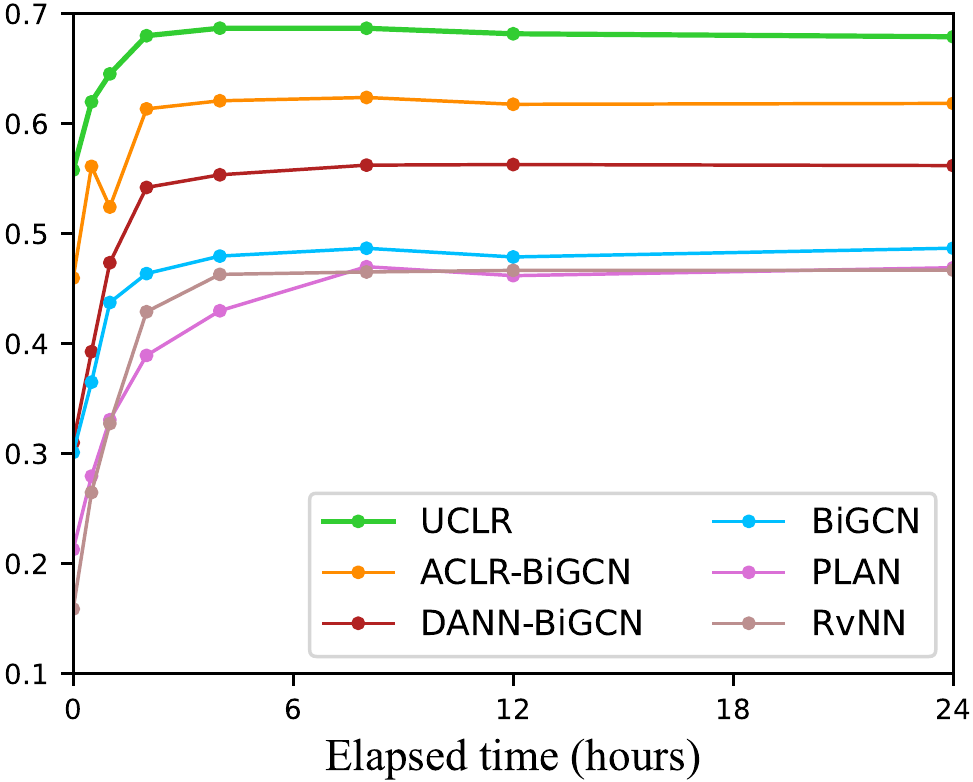}}
\label{fig:early_c}
\end{minipage}%
}%
\\
\subfigure[Chinese-COVID19 (posts count)]{
\begin{minipage}[t]{0.33\linewidth}
\centering
\scalebox{0.85}{\includegraphics[width=6cm]{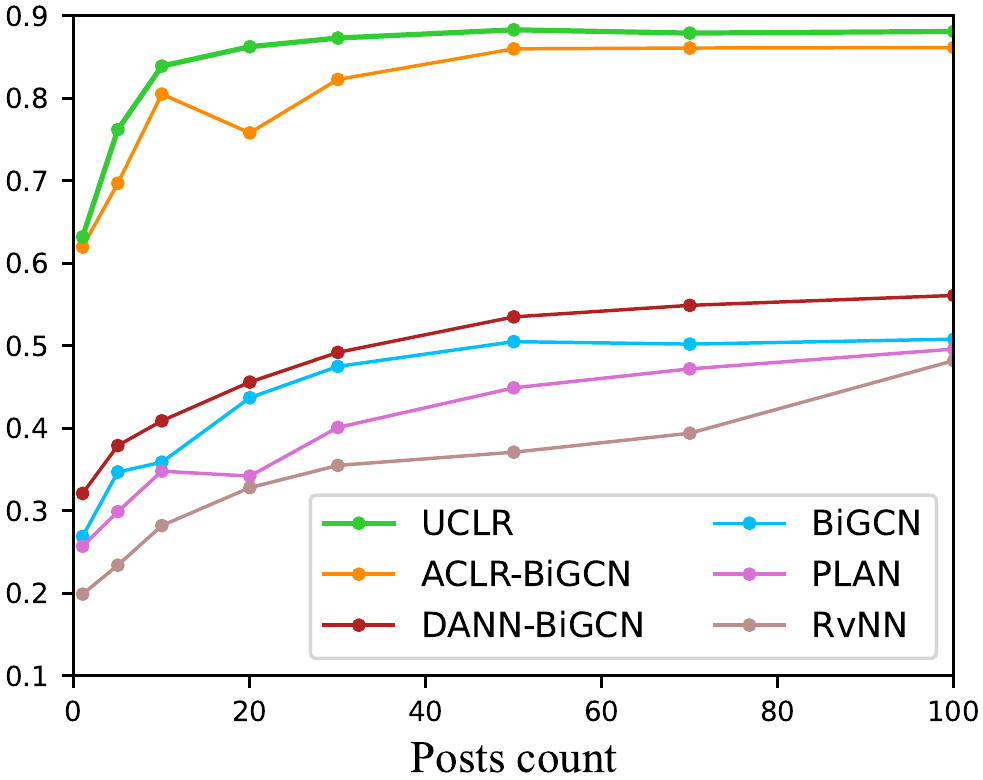}}
\label{fig:early_d}
\end{minipage}%
}%
\subfigure[Cantonese-COVID19 (posts count)]{
\begin{minipage}[t]{0.33\linewidth}
\centering
\scalebox{0.85}{\includegraphics[width=6cm]{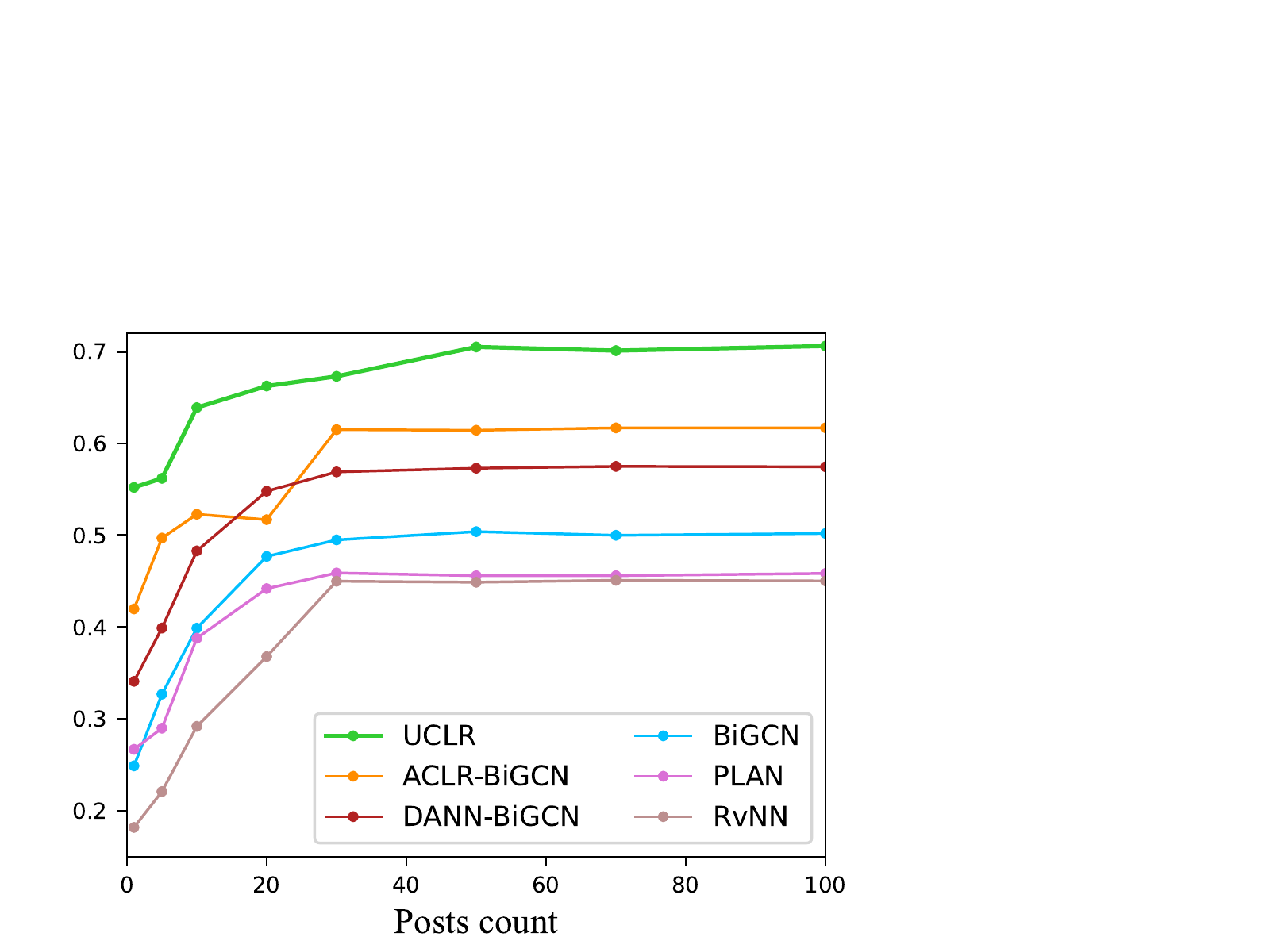}}
\label{fig:early_e}
\end{minipage}%
}%
\subfigure[Arabic-COVID19 (posts count)]{
\begin{minipage}[t]{0.33\linewidth}
\centering
\scalebox{0.85}{\includegraphics[width=6cm]{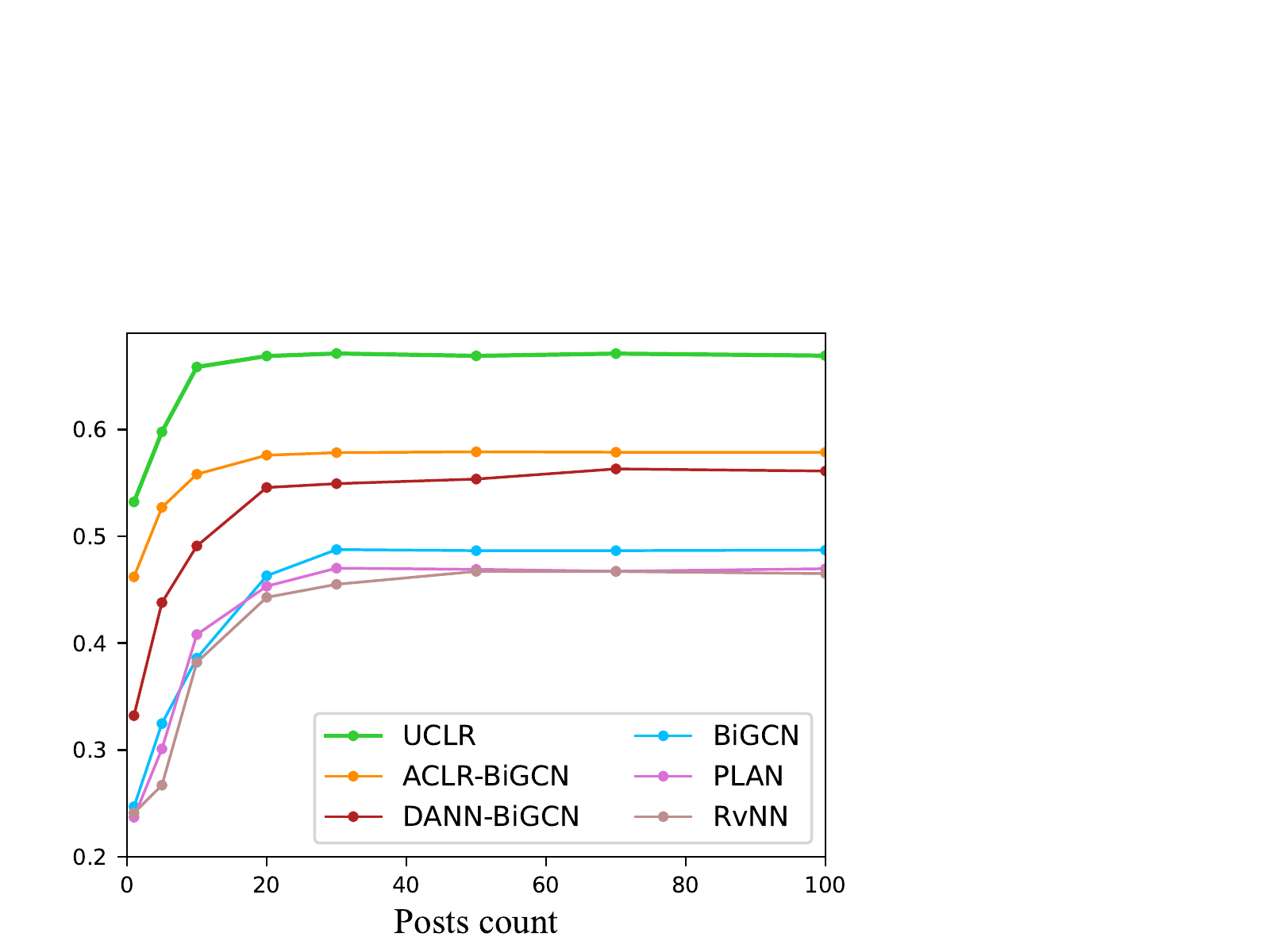}}
\label{fig:early_f}
\end{minipage}%
}%
\centering
\caption{Early detection performance Macro F1 at different checkpoints of elapsed time (or posts count).}
\label{fig:early_detection}
\end{figure*}}

\textbf{Effect of Well-resourced Data} As demonstrated in Table~\ref{ablative_1} and Table~\ref{ablative_2}, the first group shows the results for the best-performed data-driven baseline BiGCN. We observe that the model performs best if pre-trained on source data and then fine-tuned on target training data (i.e., BiGCN(S,T)), compared with the poor performance when trained on either minor labeled target data only (i.e., BiGCN(T)) or well-resourced source data (i.e., BiGCN(S)). This suggests that our hypothesis of leveraging well-resourced source data to improve the low-resource rumor detection on target data is feasible. 

\textbf{Effect of Feature Alignment} In the second group, the DANN-based model makes better use of the source data to extract domain-agnostic features, which further leads to performance improvement. Our proposed domain-Adaptive Contrastive Learning approach ACLR-BiGCN has already achieved outstanding performance compared with other baselines, which illustrates its effectiveness on domain and language adaptation. 

\textbf{Effect of Target-wise Uniform Distribution} We further notice that our UCLR-BiGCN consistently outperforms all baselines and improves the prediction performance of ACLR-BiGCN, suggesting that training model to preserve more rumor-indicative information on target data with more uniform distribution, could provide robust generalization for more accurate rumor predictions, especially in low-resource regimes.

\textbf{Effect of Multi-scale GCNs} Our proposed UCLR frameworks with Multi-scale GCNs in the third group generally perform better than the UCLR-BiGCN, which indicates the potential of Multi-scale GCNs as the backbone of our few-shot transfer framework for propagation structure representation learning, complementary to the proposed unified contrastive training paradigm.

\textbf{Effect of Data Augmentation Strategies} In the third group, we explore the effectiveness of different augmentation strategies to our proposed Target-wise Contrastive Learning. We can observe that the Graph Dropedge we employ as the data augmentation for the propagation structure is the most effective strategy, outperforming Adversarial Attacks and Feature Dropout. This is probably because Grpah Dropedge is more related to our propagation-aware framework since they are directly operated on the Event-structural scale of Multi-scale GCNs and change the structure of the propagation to produce hard examples.

\subsection{Early Detection}
Early alerts of rumors are essential to minimize their social harm. 
By setting detection checkpoints of ``delays" that can be either the count of reply posts or the time elapsed since the first posting, only contents posted no later than the checkpoints is available for model evaluation. The performance is evaluated by Macro F1 obtained at each checkpoint. To satisfy each checkpoint, we incrementally scan test data in order of time until the target time delay or post volume is reached.

Figure~\ref{fig:early_detection} shows the performances of our approach versus ACLR-BiGCN~\cite{lin2022detect}, DANN-BiGCN~\cite{ganin2016domain}, BiGCN~\cite{bian2020rumor}, PLAN~\cite{khoo2020interpretable}, and RvNN~\cite{ma2018rumor} at various deadlines. 

Firstly, we observe that the accuracies of all systems obviously increase with elapsed time or post counts, our proposed UCLR approach outperforms other counterparts and baselines throughout the whole lifecycle, which grows more quickly to supersede the other baselines and reaches a relatively high Macro F1 score at a very early period after the initial broadcast. One interesting phenomenon is that the early performance of some methods may fluctuate more or less. It is because with the propagation of the claim, there is more semantic and structural information but the noisy information is increased simultaneously. Our method only needs about 50 posts and around 4 hours with TWITTER and WEIBO as source data, respectively, to achieve the saturated performance, indicating the remarkably superior early detection performance of our method.

{
\begin{figure}[t]
\centering
\subfigure{
\begin{minipage}[t]{0.5\linewidth}
\centering
\scalebox{0.8}{\includegraphics[width=5cm]{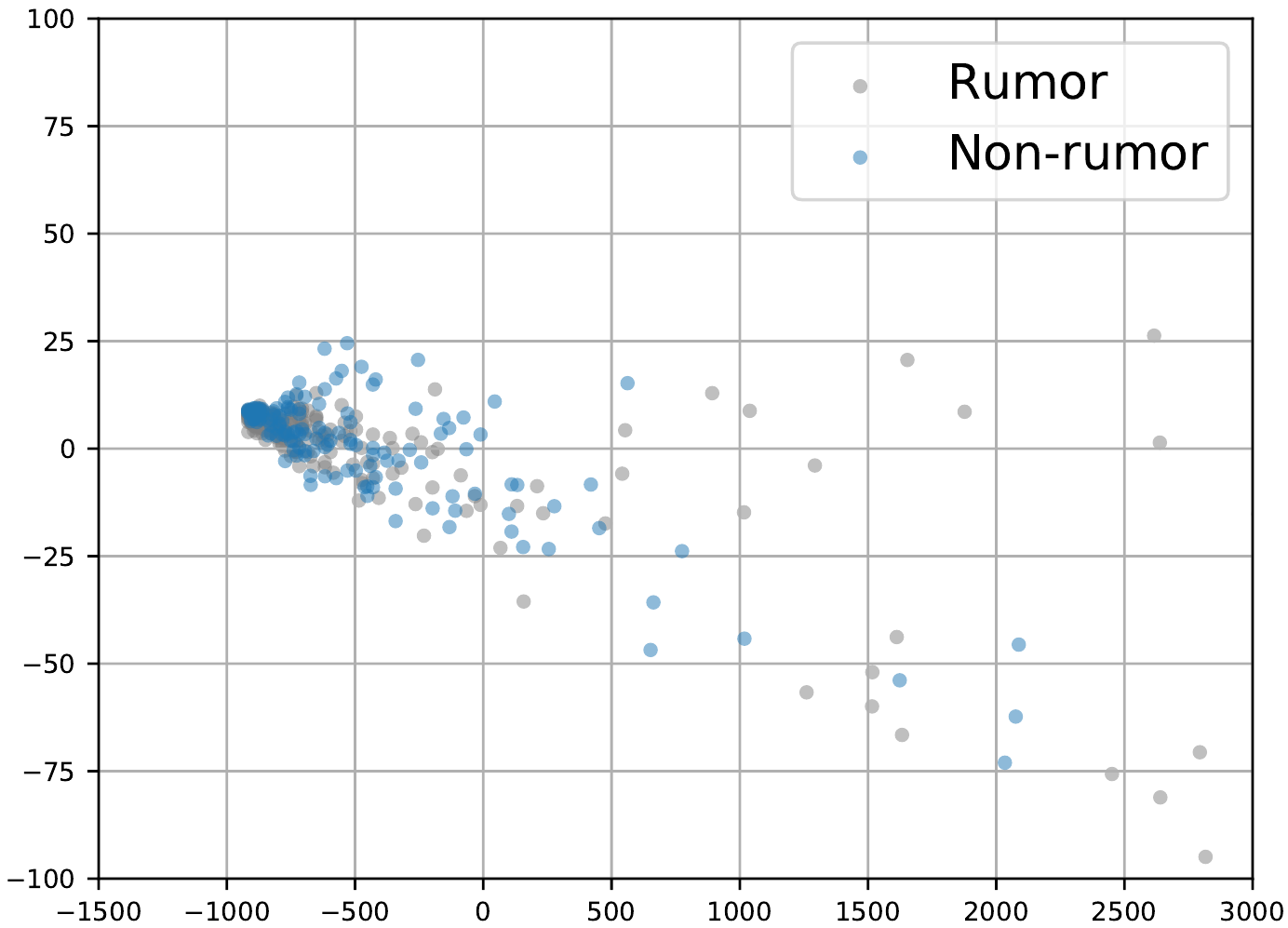}}
\label{fig:baseline_distribution}
\end{minipage}%
}%
\subfigure{
\begin{minipage}[t]{0.5\linewidth}
\centering
\scalebox{0.8}{\includegraphics[width=5cm]{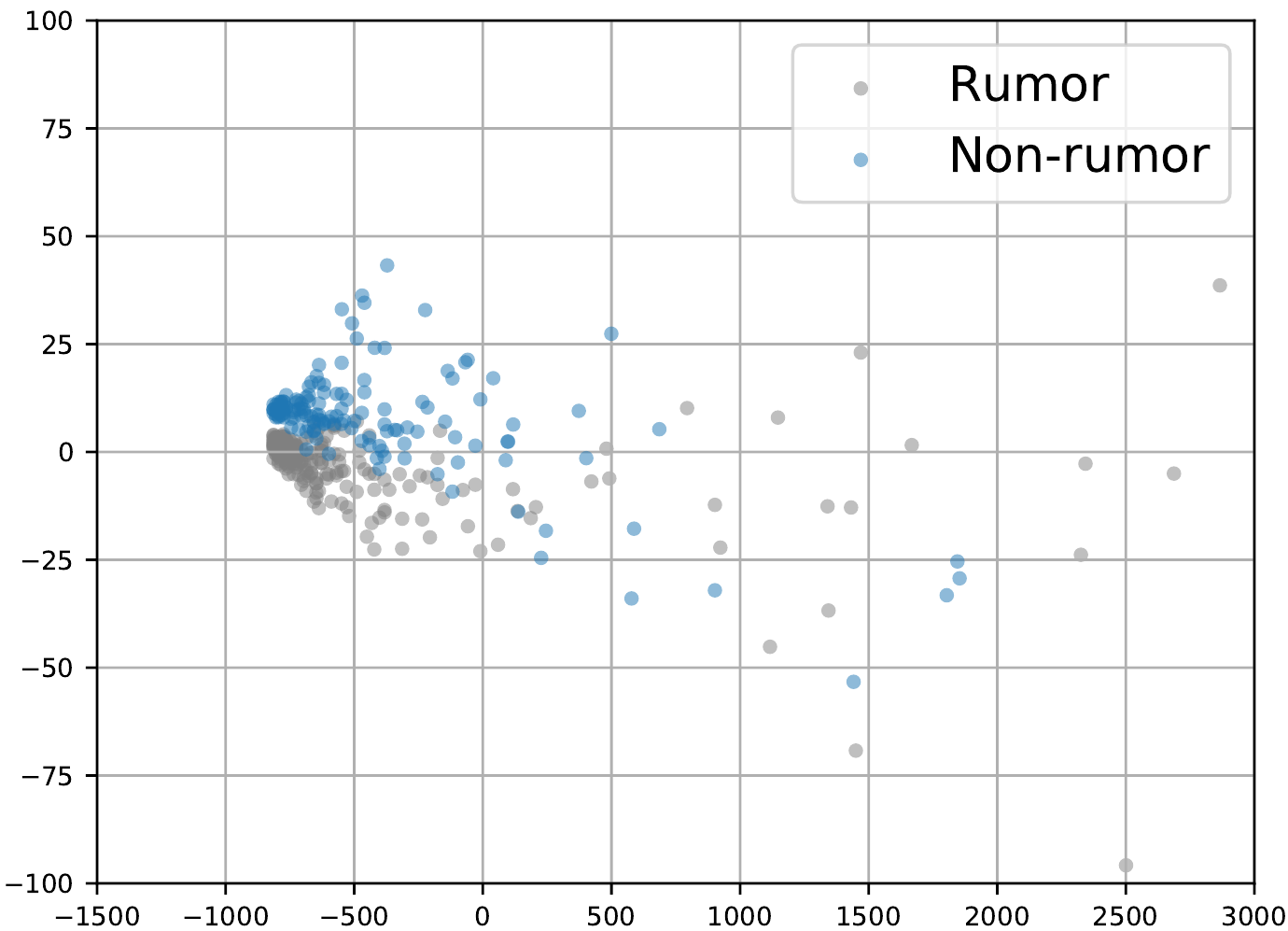}}
\label{fig:ACL_distribution}
\end{minipage}%
}%
\centering
\caption{Visualization of target event-level representation distribution for traditional classification (left) and ACLR (right) paradigms on ChiCovid data.}
\label{fig:vis}
\end{figure}}

{
\begin{figure}[t]
\centering
\subfigure{
\begin{minipage}[t]{0.5\linewidth}
\centering
\scalebox{0.8}{\includegraphics[width=5cm]{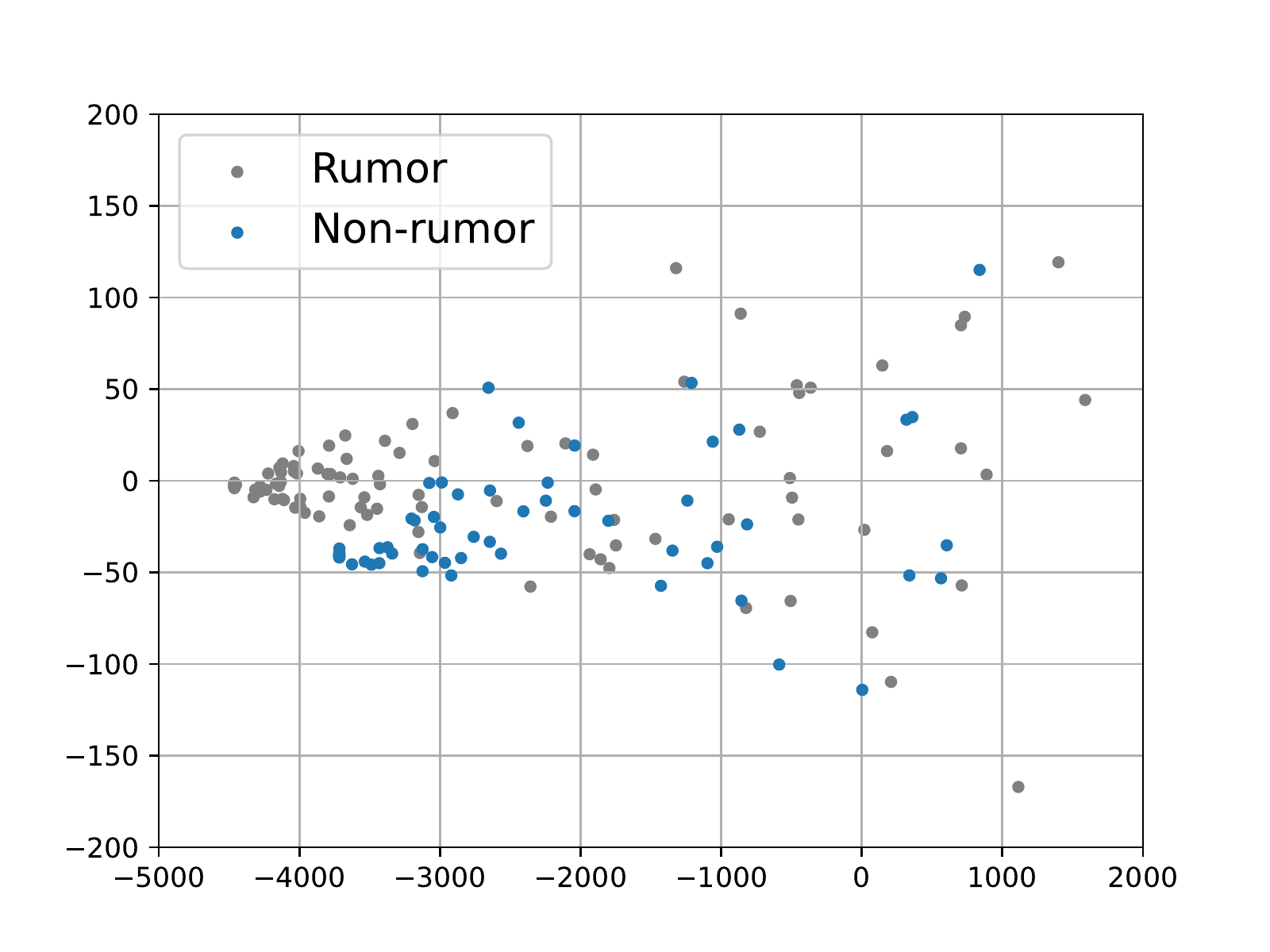}}
\label{fig:ACLR_distribution}
\end{minipage}%
}%
\subfigure{
\begin{minipage}[t]{0.5\linewidth}
\centering
\scalebox{0.8}{\includegraphics[width=5cm]{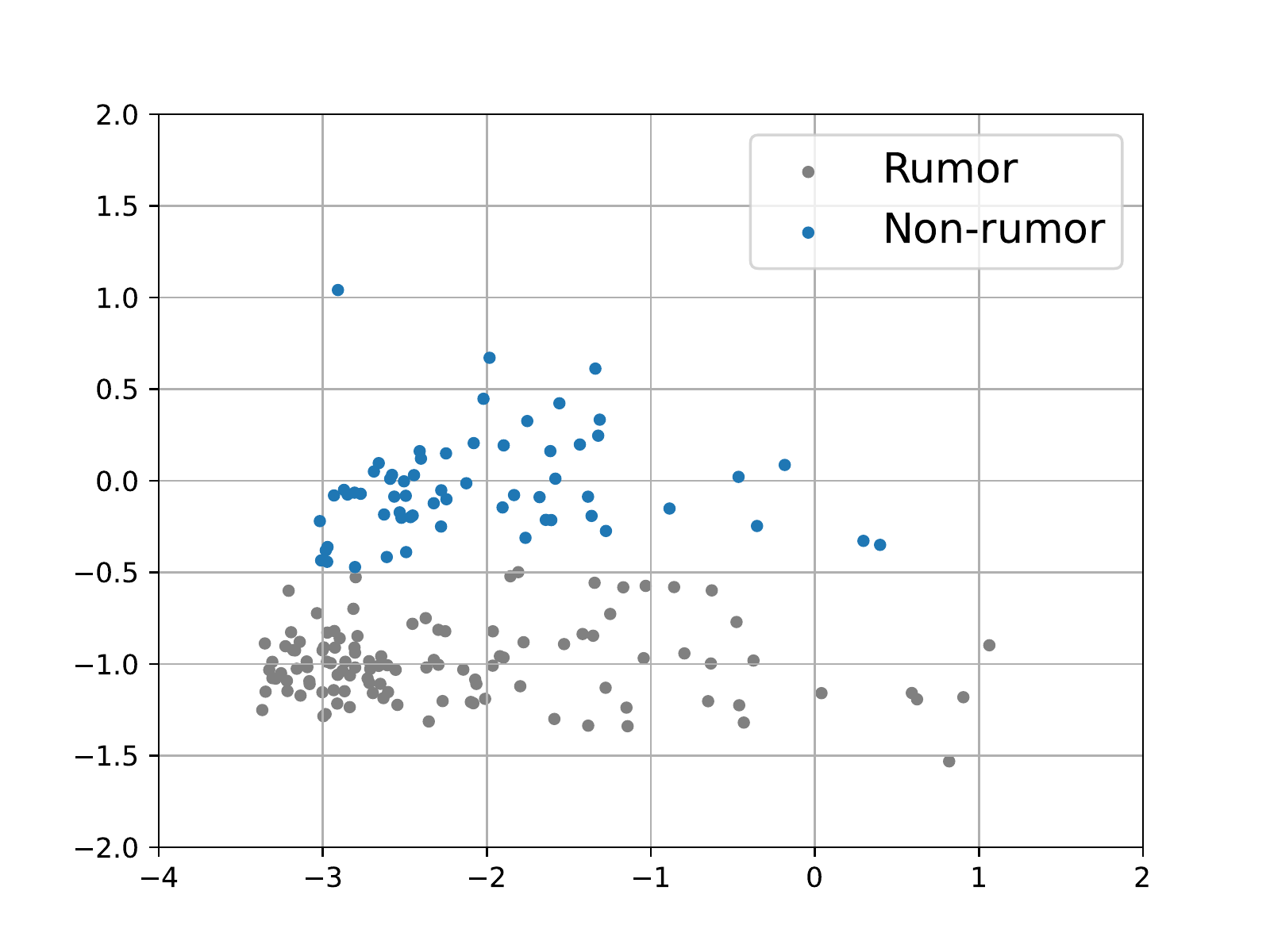}}
\label{fig:UCLR_distribution}
\end{minipage}%
}%
\centering
\caption{Visualization of target event-level representation distribution for ACLR (left) and UCLR (right) paradigms on AraCovid data.}
\label{fig:vis_}
\end{figure}}

\begin{figure*}[t]
\centering
\scalebox{0.75}{\includegraphics[width=20cm]{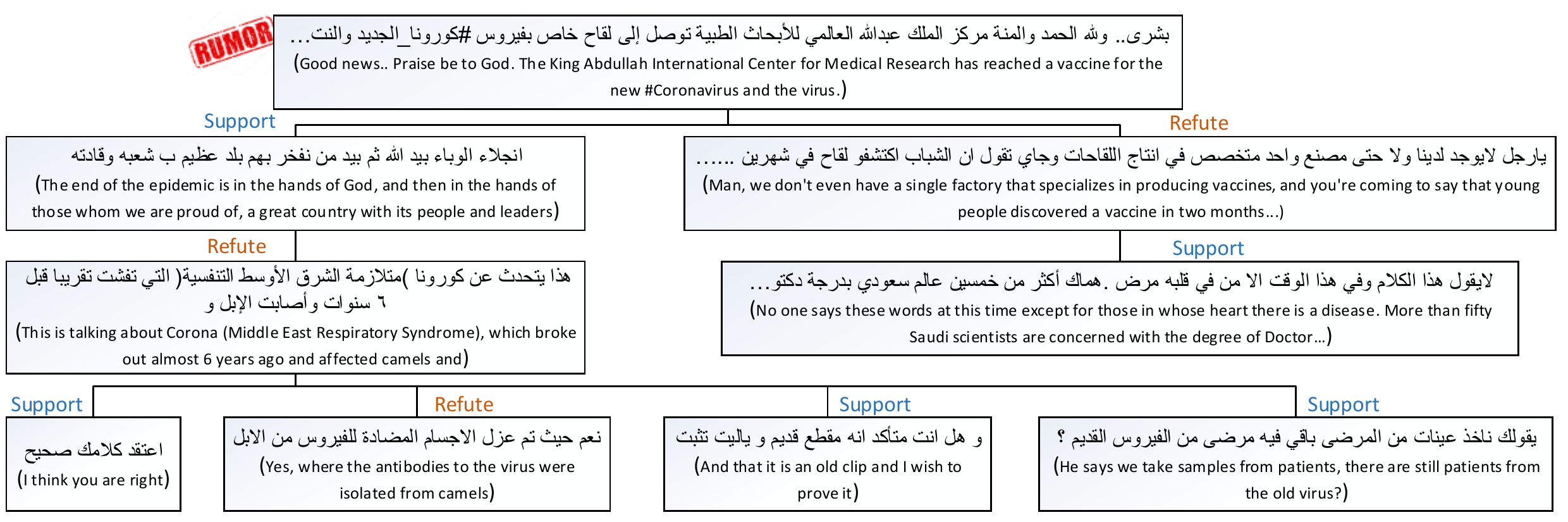}}
\caption{A sample case of correctly detected rumors of our model. We show important tweets in the propagation structure and truncate others.}
\label{fig:case}
\end{figure*}

\subsection{Feature Visualization}

Figure~\ref{fig:vis} shows the PCA visualization of learned target event-level features obtained from traditional classification (left) and ACLR (right) paradigms on Chinese-COVID19 data for analysis. The left figure represents model training with only classification loss, and the right figure uses our proposed domain-Adaptive Contrastive Learning for training. We observe that (1) due to the lack of sufficient training data, the features extracted with the traditional training paradigm are entangled, making it difficult to detect rumors in low-resource regimes; and (2) our ACLR-based approach learns more discriminative representations to improve low-resource rumor classification, reaffirming that our training paradigm can effectively transfer knowledge to bridge the gap between source and target data distribution resulting from different domains and languages. Furthermore, Figure~\ref{fig:vis_} illustrates the difference in feature visualization obtained from ACLR (left) and UCLR (right) paradigms on Arabic-COVID19 data. It is observed that, besides the better decoupling for different rumor-related labels, the UCLR achieves a relatively more evenly distributed feature set for the target data compared with the ACLR, which indicates the effectiveness of Target-wise Contrastive Learning in contributing to the generalization ability of our framework in low-resource regimes.

\subsection{Case Study of Propagation Structure}

For a more comprehensive analysis of the propagation structure, we present an example of correctly detected rumors with part of its propagation structure. The visualization of tweets in Figure~\ref{fig:case} shows that when a post challenges a rumor, it tends to elicit supportive or affirming replies that confirm the denial. Conversely, when a post endorses a rumor, it tends to trigger denials in its replies. Furthermore, it is observed that a reply typically responds to its immediate parent node rather than directly to the root claim. This observation aligns with our motivation to explore the propagation structure of rumors for representation learning. By adopting an undirected topology, the structure can be naturally modeled to capture the signals indicative of rumors and enhance representation by fully aggregating features from all informative neighbors. This enables the adaptive propagation of information association between nodes in the conversation thread along responsive parent-child relationships.

Furthermore, we can observe that the informative posts should be developed and extended around the content of the claim, i.e., the potential and implicit target to be checked. This highlights the significance of the claim content to catch informative posts. Our proposed multi-scale GCNs could integrate claim information from the claim-semantic scale with the propagation thread from the event-structural scale, to enrich the semantic context of replies and better guard the consistency of topics for the correct prediction.

{
\begin{figure}[t]
    \centering
    \resizebox{0.45\textwidth}{!}{\includegraphics{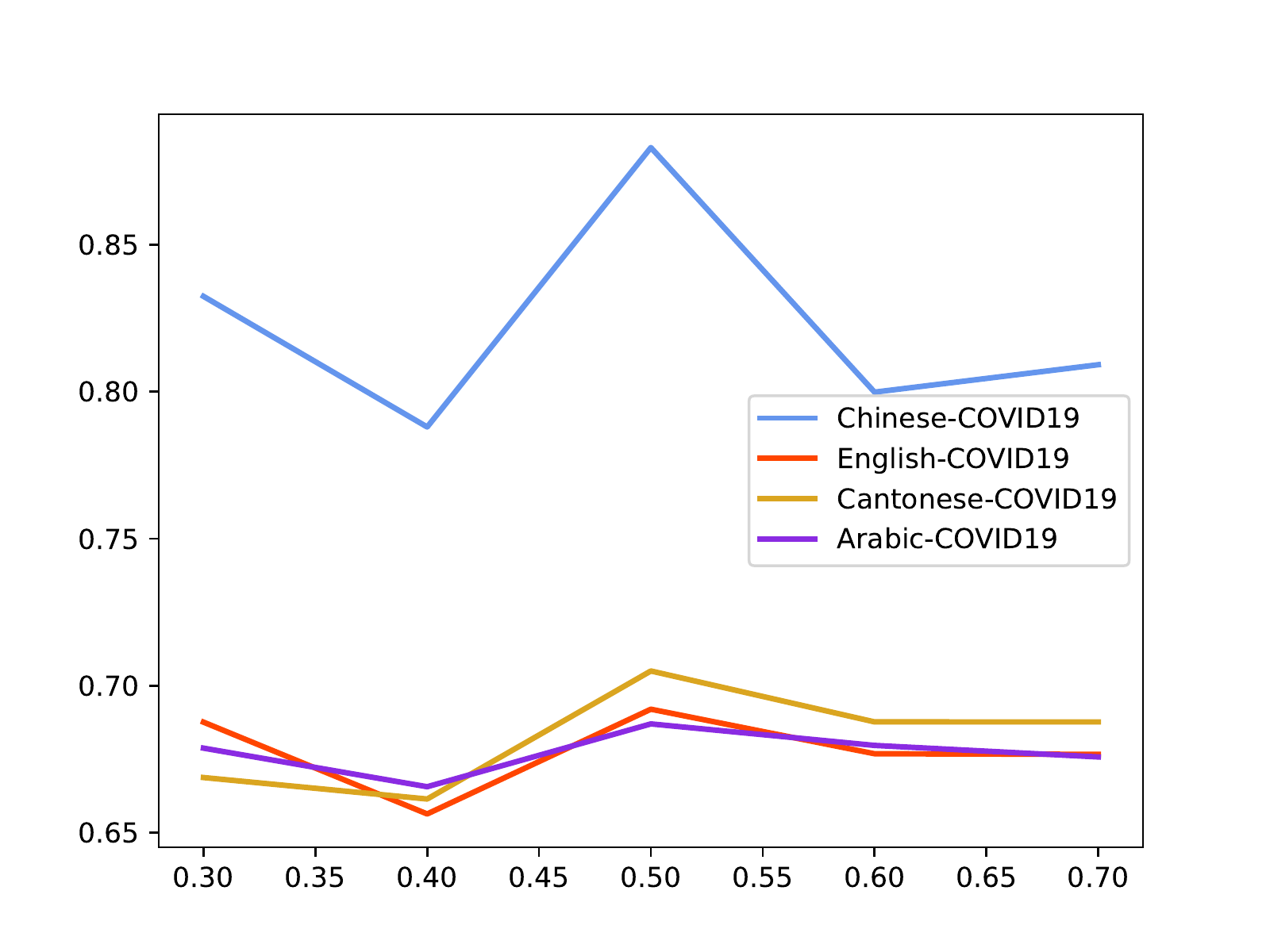}}
    \caption{Effect of trade-off hyper-parameter $\alpha$ between Classification and Contrastive Objectives.}
    \label{fig:alpha}
\end{figure}}

\subsection{Effect of Trade-off Hyper-parameter}
To study the effects on performance (Macro F1 score) of the trade-off hyper-parameter in our training paradigm, we conduct qualitative analysis under UCLR architecture (Figure~\ref{fig:alpha}). For the target data Chinese-COVID19 and Cantonese-COVID19, we use TWITTER as the source data; in terms of English-COVID19 and Arabic-COVID19, we use WEIBO as the source data. Since the platform for collecting Chinese-COVID19 data is Sina Weibo while the platform for the other three datasets is Twitter, there will be a large gap between the model's performance on Chinese-COVID19 data and its performance on the other three datasets. We can see that $\alpha = 0.5$ achieves the best performance while the point where $\alpha =0.3$ also has good performance. Looking at the overall trend, the performance fluctuates more or less as the value of $\alpha$ grows. We conjecture that this is because the unified contrastive objective, while optimizing the representation distribution, compromises the mapping relationship with labels. Such a multi-task paradigm means optimizing the traditional classification loss and the unified contrastive loss simultaneously. This setting leads to mutual interference between two tasks, which affects the convergence effect. This phenomenon points out the direction for our further research in the future.

\subsection{Discussion about Low-Resource Settings}
In this section, we evaluate our proposed framework with mono-lingual and cross-lingual source datasets to discuss the low-resource settings in our experiments. Considering the cross-domain and cross-lingual settings in Table~\ref{tab:main_results_1} of the main experiments, we also conduct an experiment in cross-domain and mono-lingual settings. Specifically, for the Chinese-COVID19 as the target data, we utilize the WEIBO dataset as the source data with rich annotations. In terms of English-COVID19, we set the TWITTER dataset as the source data. Table~\ref{tab:low-resource} depicted the results in different low-resource settings. It can be seen from the results that our model performs generally better in cross-domain and cross-lingual settings concurrently than that only in cross-domain settings, which demonstrates the key insight to bridging the low-resource gap is to relieve the limitation imposed by the specific language resource dependency besides the specific domain. Our proposed unified propagation-aware contrastive transfer framework could alleviate the low-resource issue of rumor detection as well as reduce the heavy reliance on datasets annotated with specific domain and language knowledge.

\begin{table}[t]
\centering
\caption{Rumor detection results of our proposed framework in different low-resource settings. Cross-D\&L denotes the cross-domain and cross-lingual settings and Cross-D denotes the cross-domain and mono-lingual settings.}
\label{tab:low-resource}
\resizebox{0.46\textwidth}{!}{
\begin{tabular}{l||cc||cc}
\hline
Target     & \multicolumn{2}{c|}{Chinese-COVID19} & \multicolumn{2}{c}{English-COVID19} \\ \hline
Settings   & Acc.             & Mac-$\emph{F}_1$          & Acc.             & Mac-$\emph{F}_1$           \\ \hline \hline
Cross-D\&L & 0.895            & \textbf{0.883}  & \textbf{0.773}   & \textbf{0.692}   \\ \hline
Cross-D    & \textbf{0.899}   & 0.864             & 0.752            & 0.645            \\ \hline
\end{tabular}}
\end{table}

\section{Conclusion and Future Work}
In this work, we propose a unified contrastive transfer framework with propagation structure to bridge low-resource gaps for rumor detection on social media by adapting social contextual features learned from well-resourced data to that of the low-resource breaking events. For the novel Low-Resource Rumor Detection task, our domain-adaptive contrastive learning aligns identical features from different domains and/or languages. Furthermore, we propose target-wise contrastive learning with three data augmentation strategies to optimize representations of target data more uniformly by distinguishing individual target training samples, for better generalization to unseen target data. Results on four real-world datasets show that: 1) our method is more effective and robust compared with state-of-the-art baselines; and 2) our extended unified contrastive transfer framework with target-wise contrastive learning makes further improvements over the original domain-adaptive contrastive model. We also compare different data augmentation strategies for target-wise contrastive learning and provide fine-grained analysis for interpreting how our approach works. 

For future work, we will explore the following directions: 1) We are going to explore the pre-training method with contrastive learning and then finetune the model with classification loss, which may further improve the performance and stability of the model; 2) Besides the textual information of the relevant posts, we will incorporate more information types (e.g., user profiles, post time, etc.) for improving our unified contrastive training paradigm; 3) Considering that our model has explicitly overcome the restriction of both domain and language usage in different datasets, we plan to leverage more datasets with rich annotation and apply our model to other domains and minority languages. 


%





\ifCLASSOPTIONcaptionsoff
  \newpage
\fi

\bibliography{custom}
\bibliographystyle{IEEEbib}

\end{document}